\documentclass[12pt]{article}
\usepackage[utf8]{inputenc}
\usepackage[T1]{fontenc}
\usepackage{amsmath,amssymb}
\usepackage{graphicx}
\usepackage{multirow}
\usepackage{booktabs}
\usepackage{adjustbox}
\usepackage{array}
\usepackage{longtable}
\usepackage{hyperref}
\usepackage{geometry}
\usepackage{fancyhdr}
\usepackage{setspace}
\usepackage[numbers]{natbib}
\usepackage{caption}
\usepackage{float}
\usepackage{subcaption}
\usepackage{tabularx}
\usepackage{makecell}
\usepackage{array, tabularx, booktabs, makecell}
\newcolumntype{Y}{>{\raggedright\arraybackslash}X}

\begin{document}
\thispagestyle{fancy}
\begin{center}
    {\LARGE \textbf{CheXthought: A global multimodal dataset of clinical chain-of-thought reasoning and visual attention for chest X-ray interpretation}\par}
    \vspace{1em}
{\large
Sonali Sharma$^{1,2}$, Jin Long$^{3}$, George Shih$^{4}$, Sarah Eid$^{5}$, \\
Christian Bluethgen$^{1,2}$, Francine L. Jacobson$^{5}$, Emily B. Tsai$^{2}$,\\ *Global Radiology Consortium, \\Ahmed M. Alaa$^{6,7}$, Curtis P. Langlotz$^{1,2,8}$\par
}
    \vspace{0.7em}
    {\footnotesize
    $^{1}$Center for Artificial Intelligence in Medicine and Imaging, Stanford University, Stanford, CA, USA\par
    $^{2}$Department of Radiology, Stanford University, Stanford, CA, USA\par
    $^{3}$Department of Pediatrics, Stanford University School of Medicine, Stanford, California, USA\par
    $^{4}$Department of Radiology, Weill Cornell Medicine, New York, NY, USA\par
    $^{5}$Department of Radiology, Brigham and Women's Hospital, Boston, MA, USA\par
    $^{6}$University of California, Berkeley, CA, USA\par
    $^{7}$University of California, San Francisco, CA, USA\par
    $^{8}$Department of Biomedical Data Science, Stanford University, Palo Alto, CA, USA
    }
\vspace{0.3em}

{\footnotesize \href{mailto:sonali3@stanford.edu}{sonali3@stanford.edu}}
\end{center}

\vspace{0em}
\begin{center}
\textbf{*Global Radiology Consortium}
\end{center}

{\large
Grant MacKinnon, Mustafa Ege Seker, Stephanie Nougaret, Suvrankar Datta, Yoojin Nam, Seong Ho Park, Yousif Al-Naser, Aram Can Yuce, Raif Yarol, Emre Ruhat Avci, Furkan Uncuoglu, Ekin Tunga, Mushtaq Omid, Onur Taydas, Rabia Aksoylu, Burak Yangoz, Kaan Kucukdere, Esra Aktufan, Gizem Karagol, Zeynep Nilufer Tekin, Zeynep Kazci, Ahmet Kasim Karabulut, Caner Öztürk, Fatma Can, Eren Çamur, Gorkem Durak, Yasin Karkas, Cleiton Aquino de Freitas, Klenam Dzefi-Tettey, Delaedem Tamakloe, Eyram Mawulolo Dagadu, Gideon Gawu-Mensah, George Hadjidekov, Mohamad Monif Assker, Munther Abduljalil, Hwajin Lee, Dong Yeong Kim, Jeong Min Song, Songsoo Kim, Seunghyun Song, Hee Sang Oh, Myung-Jean Lee, Ji Won Jeong, Nuri Saraç, Onat Guner, Ismet Cayir, Alexander V. Ortiz, Calvin Qian, Arne Michalson, Munkhbaatar Dagvasumberel, Ali Canturk, Betul Uzunoglu, Abidin Kılınçer, Ayşe Arı, Selman Mert, Elif Kamalak, Ahmet Baytok, Gulsum Kilickap, Esra Yurduseven, Yunus Ysr, Gurhan Abdi, Nurper Denizoğlu, Yasin Celal Gunes, Lili Koteterova, Ainara Darbayeva, Liliana De Giorgi, Giulia Ronca, Francesca Fiorella, Ersilia Salsano, Antonio Salerno, Antonello L.L. Marcigliano, Donato Bressi, Antonio Rizzo, Laura Sommella, Maria Gerardina Freda, Eugenia Amelia Lettieri, Renato Cuocolo, Andrea Ponsiglione, Kursat Karaman, Nur Betul Unal Ozdemir, Elif Yaşar, Elizabeth Zamora Endara, Jasmin Maria-Theresia Happe, Stefanus Imanuel Setiawan, Sunghyun Jung, Won Jung Kim, Andrew Refalo, Simon P. Woodbridge, Naimish Adroja, Sabeeh Syed, Faysal Al-Ma'ayah, Faris Haddad, Mikel Isla-Jover, Martin Santamaria Boado, Prashansa Gupta, Min-Fang Chao, Cheng-Hsuan Juan, Meltem Emine, Carola Brussaard, Calvin T. John, Fırat Atak, Fatih Erdem, Asli Tanrivermis Sayit, Gülten Taşkın, Busra Has, Mohamed R. Nouh, Roqiah Fatmawati, Abdul Kadir, Shyamal Vijai K, Vlastimil Valek Jr., Erencan Karakoç, Chloe DesRoche, Monish Ahluwalia, Jane Burns, Sasa Golubovic, Şerife Nur Ulusan, Mariola Gutiérrez Gallardo, Zaid Hussain Khan, Yohan Joo, Devyani Singh, Khushboo Jain, Alessandra Farchione, Lukas Müller, Maximilian Moos, VenkataGanesan Ponnalagu, Vuk Matić, Ebru Hasbay, Ryeong Ah Kim, Atinuke M Agunloye, Dhananjay Sahastrabudhe, Warissara Jutidamrongphan, Deepak Justine Viswanathan, Wen-yuan Yang, Anurag Goyal, JanJan Chai, Avnish D. Goyani, Paulina Kalinowska, Enes Deger, Joseph Ortolani, Lieve Morbée, Martin T Halicek, Mohammed Shakeebuddin Kashif, Reece J. Goiffon, Sanjana Hebbar, José Ignacio Debes, Clemente García-Hidalgo, Jamal Bani Issa, Aishwarya Yogiraj Gosavi, Kevin Anton, Tejas Phaterpekar, Ricky Hu, Navjit Singh, Asma Al Hatmi,  Jason Yao, Courtney Cheng, Tahreem Anwar, Yuki Arita, Rania Refaat, Shreyas Reddy Kankara, Suyash Gunjal.
}
\normalsize
\clearpage

\normalsize
\begin{abstract}
Chest X-ray interpretation is one of the most frequently performed diagnostic tasks in medicine and a primary target for AI development, yet current vision–language models are primarily trained on datasets of paired images and reports, not the cognitive processes and visual attention that underlie clinical reasoning. Here, we present CheXthought, a global, multimodal resource containing 103,592 chain-of-thought reasoning traces and 6,609,082 synchronized visual attention annotations across 50,312 multi-read chest X-rays from 501 radiologists in 71 countries. Our analysis reveals clinical reasoning patterns in how experts deploy distinct visual search strategies, integrate clinical context, and communicate uncertainty. We demonstrate the clinical utility of CheXthought across four dimensions. First, CheXthought reasoning significantly outperforms state-of-the-art vision–language model chain-of-thought in factual accuracy and spatial grounding. Second, visual attention data used as an inference-time hint recovers missed findings and significantly reduces hallucinations. Third, models trained on CheXthought data achieve significantly higher accuracy in pathology classification, visual faithfulness, temporal reasoning and uncertainty communication. Fourth, leveraging CheXthought's multi-reader annotations, we predict both human–human and human–AI disagreement directly from an image, enabling transparent communication of case difficulty, uncertainty and model reliability. These findings establish CheXthought as a resource for advancing multimodal clinical reasoning and the development of more transparent, interpretable vision–language models.
\end{abstract}
\normalsize
\clearpage

\section{Introduction}

Chest X-ray (CXR) interpretation is one of the most commonly performed diagnostic imaging tasks across emergency, inpatient, and outpatient settings globally and a primary target for artificial intelligence (AI). As of early 2026, over 35 FDA-cleared AI devices and 10 open-source vision-language models (VLMs) targeting chest radiographs are available \citep{Raoof2012, FDA2025}. As these models are increasingly deployed in clinical settings, understanding how they arrive at a final answer is essential for patient safety, clinician trust and regulatory oversight. Reasoning models with chain-of-thought (CoT) capability have emerged as a mechanism for improving transparency, offering an interpretable window into model decision-making that supports auditing and patient safety \citep{Lievin2024}. However, previous work has demonstrated that current CoT outputs are often unfaithful, producing plausible-sounding rationales that are disconnected from the model's actual internal computations \citep{Turpin2023}, and frontier VLMs can achieve high benchmark accuracy without any image input at all, calling into question whether these models truly reason about what they see \citep{Asadi2026}.

Currently, the majority of VLMs are trained on images paired with diagnostic labels or radiology reports, neither of which characterizes the underlying expert reasoning process or the visual attention patterns that guide interpretation. In practice, clinical reasoning extends beyond pattern recognition; it requires probabilistic reasoning under uncertainty, integration of clinical context with visually grounded evidence, and the bridging of population-based knowledge to individual cases \citep{Andreoletti2019}. In radiology, this reasoning unfolds alongside directed visual attention, as radiologists systematically search across anatomical regions, prioritizing structures, revisiting areas of concern, and spatially anchoring their observations to specific image features. While a small number of prior studies have leveraged eye-tracking to approximate visual attention during image interpretation, existing datasets are limited to small numbers of readers within single institutions, lack inter-reader variability, and do not capture the entire step-by-step verbalized reasoning process that accompanies visual search \citep{Karargyris2021, Lanfredi2022}. These constraints stem largely from the logistical and financial challenges of eye-tracking itself, which requires specialized hardware with physically co-located radiologists, limiting the geographic and demographic diversity of annotator cohorts and contributing to broader biases in medical imaging datasets \citep{Kaushal2020}.

To address these challenges, we introduce CheXthought, the first large-scale global human chain-of-thought reasoning and visual attention dataset for chest X-ray interpretation. CheXthought contains 103,592 chain-of-thought reasoning traces paired with 6,609,082 synchronized visual attention coordinates across 50,312 chest X-rays drawn from the CheXpert Plus dataset \citep{CheXpertPlus}. Each reasoning trace captures step-by-step verbalized clinical reasoning while annotators simultaneously traced their visual attention across the image. Using this resource, we first analyze CheXthought data to characterize patterns in how experts clinically reason. We then perform blinded expert evaluation comparing human-authored reasoning to CoTs generated by state-of-the-art vision--language models. We further investigate whether visual attention can be used as an inference-time hint to improve pathology classification performance without retraining. Next, we systematically isolate the effect of supervision by training models on human CoT, synthetic CoT, report-only data, and both CoT and visual attention data, evaluating their impact on visual faithfulness, temporal reasoning and pathology classification. Finally, leveraging the multi-reader nature of the dataset, we train a model to predict both human--human and AI--human disagreement directly from images, providing a proxy for case complexity and diagnostic uncertainty.

\section{Results}

\begin{figure}[H]
\centering
\makebox[\textwidth][c]{\includegraphics[width=1.3\textwidth]{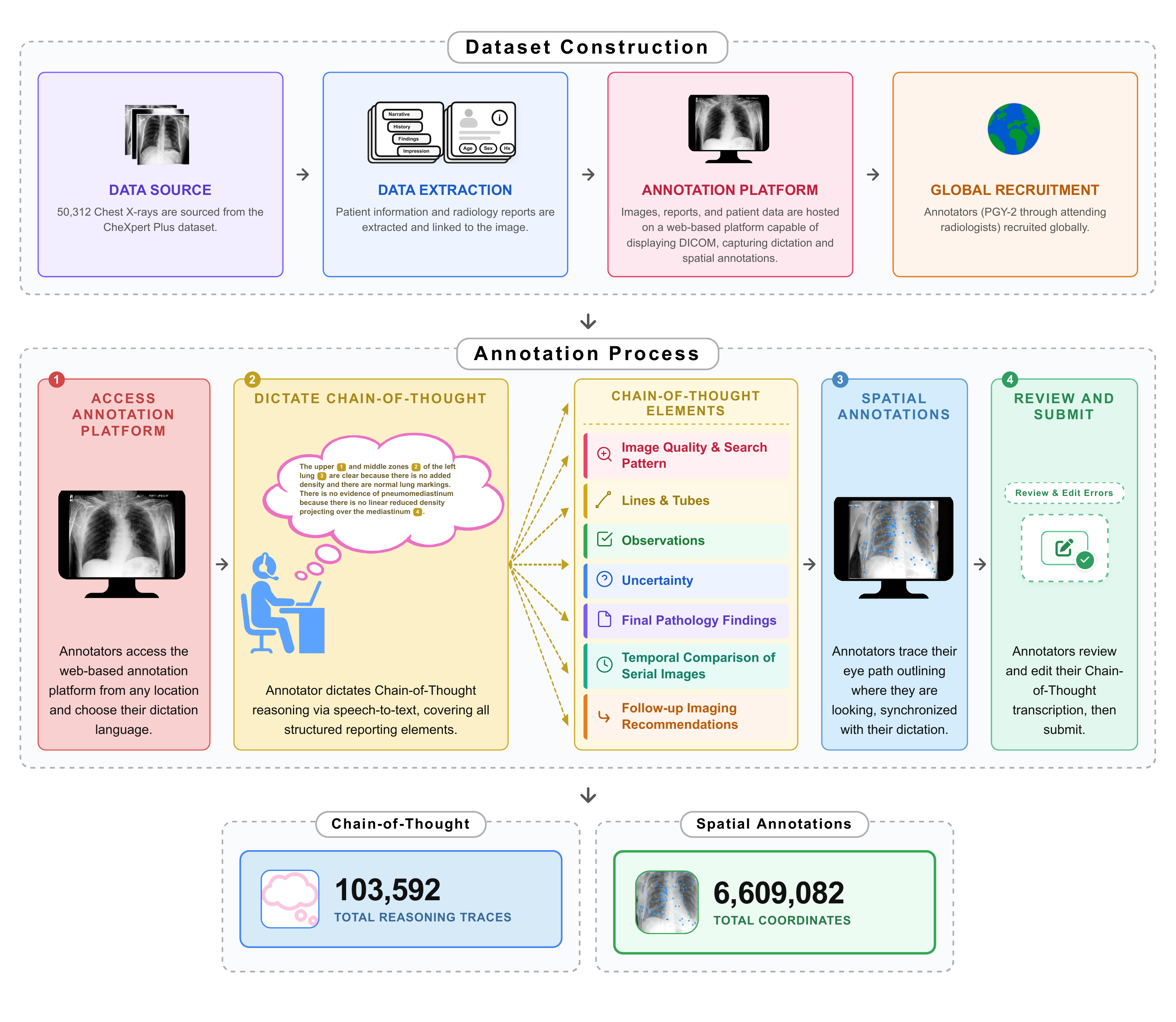}}
\caption{\textbf{(A)} CheXthought Dataset Construction and Annotation Process}
\label{fig:Fig1A}
\end{figure}

Figure~\ref{fig:Fig1A} (Panel A) summarizes the dataset construction and annotation process. To capture reasoning and visual attention simultaneously, we developed a custom, browser-based DICOM annotation platform with full PACS-style functionality, including window/level adjustment, zoom, pan, and measurement tools. Because the interface ran in a standard web browser, annotators could participate globally without local software installation or specialized hardware (Supplementary Figure~\ref{fig:Supplement1}).

Annotators were instructed to verbalize their reasoning as comprehensively as possible while simultaneously tracing image regions corresponding to their visual attention by clicking and dragging the cursor over areas of interest. In this work, we define visual attention as an explicit representation of the regions annotators actively observed as relevant while articulating their reasoning. Unlike eye-tracking data, which primarily reflects unconscious perceptual processes, this approach encodes intentional, reasoning-aligned spatial references. Cursor-based tracing of visual attention is not equivalent to eye-gaze fixation and does not capture true eye-movement dynamics; however, this approach enabled scalable, globally distributed data collection without requiring specialized hardware or controlled laboratory environments.

Visual attention was captured through an interactive overlay aligned with the DICOM viewer, allowing annotators to mark regions of interest via pointer-based tracing. Coordinates were recorded in image-relative space to remain consistent under zoom and pan operations, ensuring accurate spatial correspondence to the underlying image. Each spatial marker was indexed and synchronized with the corresponding segment of verbalized reasoning, enabling explicit linkage between textual interpretation and image-referenced evidence. Spoken input was recorded using each annotator’s local computer microphone and transcribed in real time via a continuous speech-recognition pipeline implemented directly in the browser. The transcription system was optimized for radiologic dictation by incorporating a domain-specific vocabulary and post-processing corrections for common medical terminology (e.g., “hilum,” “cardiomegaly”), and by applying overlap-aware merging to maintain a coherent, continuously growing transcript across interim and finalized recognition segments.

We captured hypotheses, visual observations, deliberations, differential diagnoses, and explicit statements of uncertainty to reflect cognitive processes rather than the compressed output typical of formal radiology reports. A structured chain-of-thought (CoT) framework was provided to guide annotators through key components of clinical reasoning while preserving their natural interpretation style without constraining content (Supplementary Table~\ref{tab:CoT}). Annotators reviewed and edited transcripts for accuracy before submission. Figure~\ref{fig:Fig1B} (Panel B) shows an example of a CheXpert Plus report paired with its corresponding CheXthought CoT, illustrating reasoning steps, uncertainty, and spatial grounding that are typically absent from standard radiology reports.

\subsection{A globally diverse annotator cohort}

A total of 501 radiologists, radiology residents, and fellows across 71 countries and all inhabited continents contributed to CheXthought, making it one of the most geographically diverse medical imaging datasets constructed to date. Annotators spanned North America (United States, Canada, Mexico); South America (Argentina, Brazil, Chile, Colombia, Ecuador, Guyana); Europe (Albania, Austria, Belarus, Belgium, Bulgaria, Czech Republic, France, Germany, Greece, Ireland, Italy, Malta, Romania, Serbia, Spain, Sweden, Switzerland, United Kingdom); Africa (Cameroon, Egypt, Ethiopia, Ghana, Kenya, Madagascar, Malawi, Morocco, Nigeria, Rwanda, Sudan, Tanzania, Uganda); Asia (Armenia, Azerbaijan, Bahrain, China, Hong Kong, India, Indonesia, Iraq, Japan, Jordan, Kazakhstan, Kuwait, Lebanon, Malaysia, Mongolia, Nepal, Oman, Pakistan, Philippines, Qatar, Saudi Arabia, Singapore, South Korea, Taiwan, Türkiye, United Arab Emirates, Uzbekistan, Vietnam); and Oceania (Australia, Fiji, New Zealand).

Eligibility was restricted to physicians at or beyond post-graduate year 2 (PGY-2) of an accredited radiology residency program. Recruitment was conducted through a global outreach effort spanning international radiology and medical imaging societies, residency programs, academic medical centers, and private-practice groups. Figure~\ref{fig:Fig1C} (Panel C) shows the geographic distribution of CoT contributions across the top 25 contributing countries.

The annotator cohort comprised PGY-2 residents (n = 44; 8.8\%), PGY-3 residents (n = 76; 15.2\%), PGY-4 residents (n = 67; 13.4\%), PGY-5 residents (n = 45; 9.0\%), fellows (n = 47; 9.4\%), and attending/staff radiologists (n = 222; 44.3\%). Gender representation included 343 men (68.5\%), 157 women (31.3\%), and 1 annotator identifying as non-binary (0.2\%). Attending/staff radiologists contributed the largest share of CoTs and spatial annotations (n = 35,739; 34.5\%), followed by PGY-3 residents (n = 19,683; 19.0\%), PGY-4 residents (n = 15,021; 14.5\%), PGY-2 residents (n = 11,913; 11.5\%), PGY-5 residents (n = 11,499; 11.1\%), and fellows (n = 9,737; 9.4\%).

\addtocounter{figure}{-1}
\begin{figure}[H]
\centering
\makebox[\textwidth][c]{\includegraphics[width=1.3\textwidth]{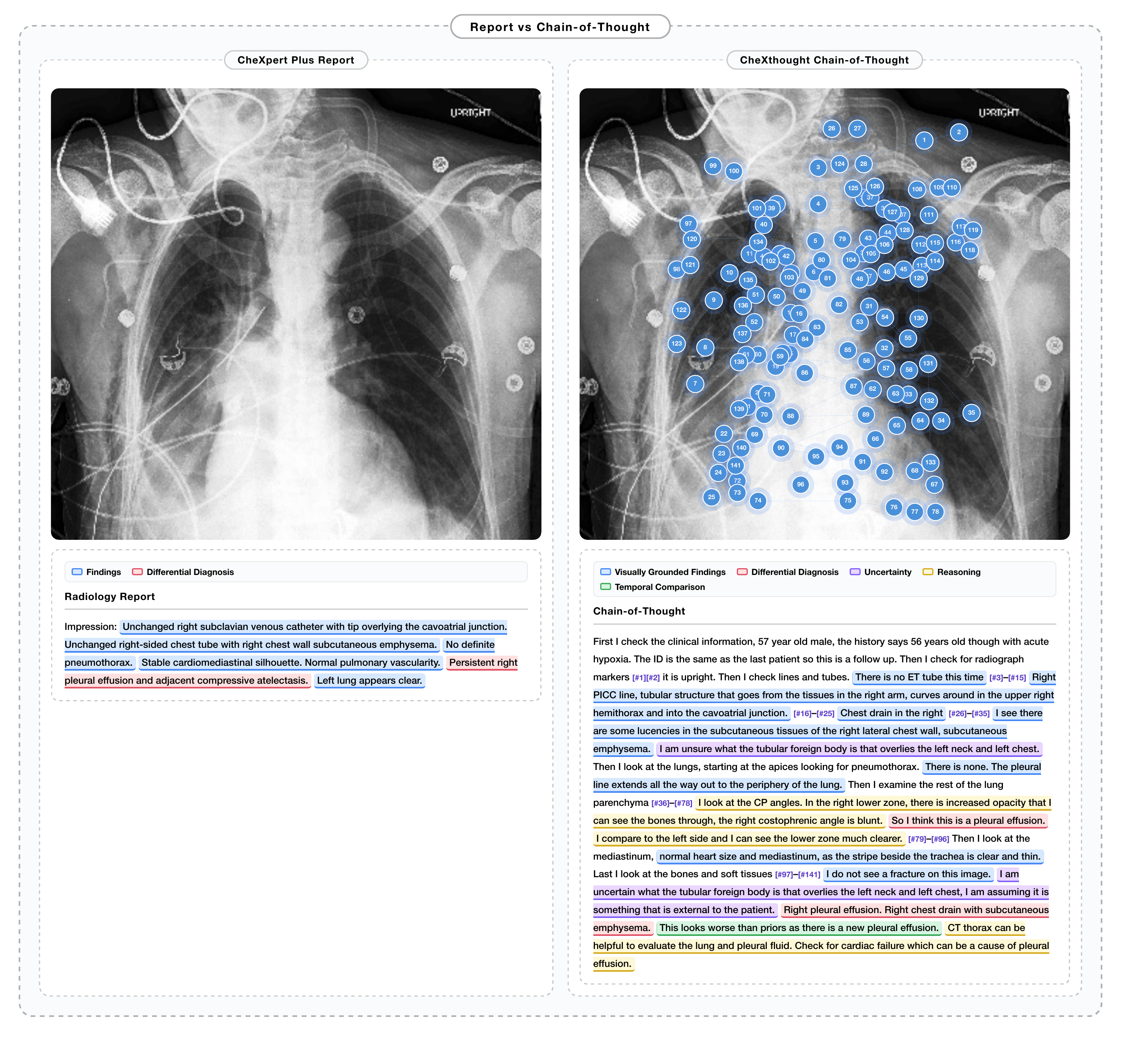}}
\caption{\textbf{(B)} Comparison Between CheXpert Plus Report and CheXthought CoT}
\label{fig:Fig1B}
\end{figure}

\addtocounter{figure}{-1}
\begin{figure}[H]
\centering
\makebox[\textwidth][c]{\includegraphics[width=1.3\textwidth]{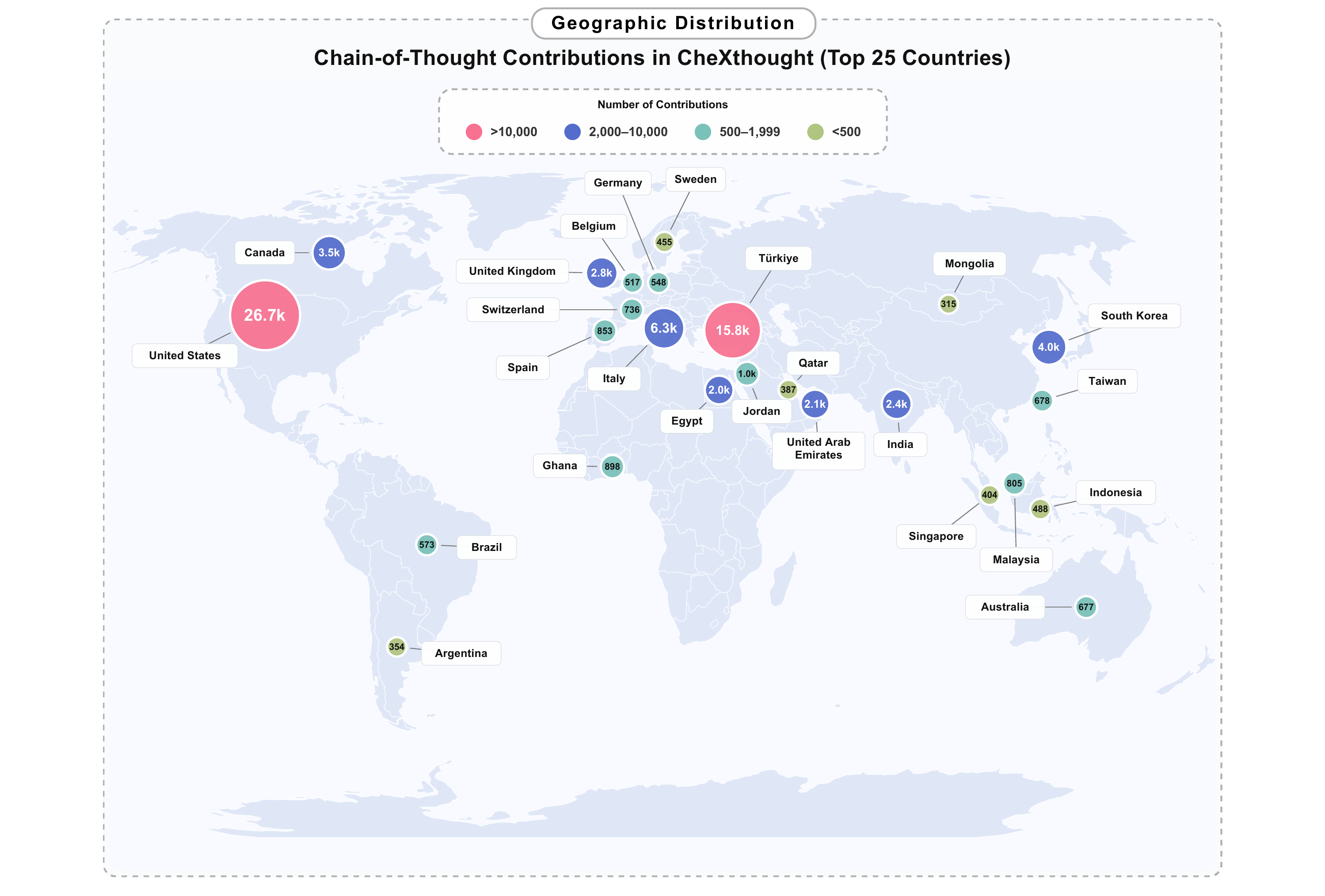}}
\caption{\textbf{(C)} Geographic Distribution of Annotators Contributing Chain-of-Thought and Visual Attention Annotations by Top 25 Countries}
\label{fig:Fig1C}
\end{figure}

\subsection{Content of chains-of-thought}

Each CoT had a median length of 117 words (IQR 67–177) and referenced a median of 66 spatial coordinates (IQR 57–73). Annotators spent a median of 5.3 minutes per CoT (mean 6.9 minutes; IQR 2.5–9.6 minutes), with modest variation across training levels (PGY-2: 5.2 min; PGY-3: 5.9 min; PGY-4: 5.6 min; PGY-5: 4.6 min; fellow: 6.0 min; attending/staff: 5.0 min). Time per CoT also varied by pathology, with shorter durations for normal chest X-rays with no findings (4.8 min) and longer durations for fracture (6.4 min) and enlarged cardiomediastinum (6.6 min). 

\subsubsection{Pathologies identified}

The full spectrum of pathologies identified across all annotator chains-of-thought can be found in Extended Data Table~\ref{tab:ExtendedData1}. A total of 59 distinct findings were extracted. The five most frequently mentioned CheXpert labels were No Finding (n = 37,637; 36.3\%), Cardiomegaly (n = 34,943; 33.7\%), Pleural Effusion (n = 22,347; 21.6\%), Edema (n = 4,783; 4.6\%), and Enlarged Cardiomediastinum (n = 4,488; 4.3\%). For the purpose of analysis, we mapped these findings to the 14 standardized CheXpert labels: No Finding, Enlarged Cardiomediastinum, Cardiomegaly, Lung Opacity, Lung Lesion, Edema, Consolidation, Pneumonia, Atelectasis, Pneumothorax, Pleural Effusion, Pleural Other, Fracture, and Support Devices~\citep{CheXpert2019}. This mapping consolidates related findings under a common taxonomy, for example, ``pulmonary edema'' and ``pulmonary vascular congestion'' map to \textit{Edema}, while ``infiltrate'' and ``reticular pattern'' map to \textit{Lung Opacity}. Similarly, device-specific mentions such as ``central venous catheter,'' ``endotracheal tube,'' and ``pacemaker/ICD'' are grouped under \textit{Support Devices}. 

\subsubsection{Temporal annotation of serial images}

The 50,312 chest radiographs in CheXthought corresponded to 45,272 distinct imaging studies from 18,046 patients, with a mean of 2.51 studies per patient (median 1; IQR 1--1; maximum 91). A total of 4,060 patients (22.5\%) had two or more distinct studies, while 13,986 patients (77.5\%) contributed a single imaging study. 87,197 (84.1\%) CoTs contained temporal references. Pathologies within CoTs were classified as improved (n = 7,691), worsened (n = 9,562), or stable (n = 85,534) and categories were not mutually exclusive.

Pleural Effusion was the most frequently mentioned pathology across all categories (improved: 1,743; worsened: 2,407; stable: 4,838). Support Devices (1,054 improved; 706 worsened; 4,692 stable) and Cardiomegaly (294; 512; 4,703) were predominantly stable, and No Finding was almost entirely stable (23,070 stable vs.\ 70 improved and 139 worsened).

Worsening was most frequent for Edema (1,448), Lung Opacity (1,332), Consolidation (643), and Pneumonia (342), whereas improvement was most frequent for Pleural Effusion (1,743), Edema (1,111), Support Devices (1,054), and Pneumothorax (635). Pneumothorax showed a notably mixed trajectory (635 improved, 551 worsened, 899 stable).

\subsection{Clinical reasoning patterns captured in CheXthought}

We analyze CheXthought data to characterize clinical reasoning patterns across both verbalized reasoning and visual attention of annotators (Figure~\ref{fig:clinical1}). Because each image in CheXthought was independently interpreted by 2 to 36 radiologists, the dataset enables direct measurement of how reasoning, visual attention, and diagnostic conclusions vary across readers.

In the sections that follow, we report two complementary forms of agreement: inter-reader agreement across independent interpretations, and concordance between each chain-of-thought (CoT) and the original CheXpert Plus report associated with the image. Accordingly, we operationalize “accuracy” as concordance with the CheXpert report, providing a standardized, large-scale reference for consistent comparison across strategies and conditions. However, we recognize that the CheXpert report reflects a single-reader interpretation that may not always be definitive ground truth. 

\subsubsection{Clinical context integration has pathology-dependent effects}

Clinical context, including patient history, indication, demographics, and references to prior studies, was documented in the associated report for 41,091 of the 50,312 images (81.7\%), corresponding to 84,515 CoTs that had context-available images (Figure~\ref{fig:clinical1} Panel A). Among these, annotators explicitly referenced clinical context during reasoning in 54.9\% of CoTs ($n = 46{,}404/84{,}515$). Referencing rates were comparable across training levels, ranging from 53.6\% (attending/staff; $n = 35{,}550$) to 56.8\% (PGY-2; $n = 10{,}009$), with intermediate values for PGY-3 (56.1\%; $n = 13{,}325$), PGY-4 (54.2\%; $n = 10{,}280$), PGY-5 (56.6\%; $n = 6{,}991$), and fellows (54.5\%; $n = 7{,}077$). Overall, CoTs that referenced clinical context achieved modestly higher accuracy than those that did not (29.4\% vs.\ 24.9\%; $\Delta = +4.5$~pp; Mann--Whitney $p < 0.0001$, Cohen's $d = 0.14$). However, the magnitude and even the direction of this effect varied substantially by pathology. Four pathologies showed a statistically significant association between context referencing and impression overlap. The effect was largest for pneumothorax ($\Delta = +16.0$~pp; $d = 0.48$; $p < 0.0001$), followed by cardiomegaly ($\Delta = +5.9$~pp; $d = 0.19$; $p < 0.0001$) and lung opacity ($\Delta = +3.9$~pp; $d = 0.12$; $p < 0.0001$). Pleural effusion reached significance with a small negative association ($\Delta = -0.8$~pp; $d = -0.03$; $p = 0.031$). The remaining ten pathologies---including consolidation, edema, pneumonia, fracture, and support devices---showed no significant benefit from context integration. These pathologies for which context provided the largest benefit---pneumothorax, cardiomegaly, and lung opacity---tend to be those for which the clinical indication frequently names or strongly implies the diagnosis (e.g., pneumothorax with post-procedural dyspnoea, cardiomegaly with known heart failure, lung opacity with suspected infection).

To control for case-level confounding---where context-available images may differ systematically in difficulty from those without---we conducted a paired image-level analysis. Among the 25,666 images read by at least one annotator who referenced context and at least one who did not, we compared accuracy within each image using a Wilcoxon signed-rank test. Context-referencing CoTs outperformed non-referencing CoTs by only $+1.38$~pp on average ($W = 44{,}809{,}156$; $p < 0.0001$; $d = 0.034$). Context helped in 29.4\% of images, hurt in 26.7\%, and made no difference in 43.9\%.

\subsubsection{Early versus later-coordinate detection of pathology}

Using a bracket-to-coordinate linking approach, in which each pathology keyword in the CoT was matched to the nearest coordinate tag \texttt{[\#N]}, we measured the coordinate index at which each finding first received visual attention (Figure~\ref{fig:clinical1} Panel B). Annotators attended to Support Devices within the first four coordinates in 65\% of cases and Pneumothorax in 45\%, whereas Edema, Pneumonia, and Lung Lesions received visual attention within the first four coordinates in only 4--5\% of cases. This ordering broadly mirrors the visual conspicuity of these findings: high-contrast, linear, or silhouette-based abnormalities tend to draw early visual attention, while diffuse or texture-based findings are attended later after broader anatomic survey.

Early visual attention did not consistently coincide with higher accuracy. Among the seven pathologies for which the association between detection timing and accuracy reached significance, six showed an inverse pattern: CoTs that first attended to the finding after the fourth coordinate achieved higher accuracy than those attending within the first four. The effect was strongest for pneumothorax (early-attention accuracy: 39.7\%; late-attention accuracy: 72.4\%; Fisher's exact $p < 0.0001$; $d = 0.35$), followed by cardiomegaly (23.0\% vs.\ 42.0\%; $p < 0.0001$; $d = 0.20$) and lung opacity (34.7\% vs.\ 49.4\%; $p < 0.0001$; $d = 0.15$). Smaller but significant effects in the same direction appeared for consolidation (40.1\% vs.\ 55.1\%; $d = 0.07$), atelectasis (47.3\% vs.\ 68.6\%; $d = 0.06$), and edema (44.5\% vs.\ 53.6\%; $d = 0.05$; all $p < 0.0001$). Pleural Effusion showed a marginal effect ($p = 0.032$; $d = -0.008$). By contrast, the remaining pathologies showed no significant association between detection timing and accuracy. Across training levels, accuracy on pathologies detected within the first four coordinates was not meaningfully different, with first-4 hit rates ranging narrowly from 19.1\% (PGY-4) to 20.3\% (Attending/Staff).

\subsection{Inter-reader variability on multi-read images}

Figure~\ref{fig:clinical1} summarizes inter-reader variability across all reliability dimensions: pathology-label agreement, agreement by training level, visual attention agreement, semantic similarity of CoTs, and agreement between CheXthought CoTs and the original CheXpert Plus reports.

\subsubsection{Inter-reader agreement on pathology findings}

Inter-reader agreement on pathology labels was quantified using Fleiss' $\kappa$, generalized to accommodate a variable number of raters per subject (2 to 36 independent reads per image; $n = 22{,}712$ multi-read images, $47{,}297$ total reads), applied separately to each of the 14 CheXpert pathologies (Figure~\ref{fig:clinical1} Panel C).

Agreement varied markedly across findings. Three pathologies with unambiguous imaging signatures reached near-perfect agreement: Pneumothorax ($\kappa = 0.92$), Pleural Effusion ($\kappa = 0.90$), and Cardiomegaly ($\kappa = 0.89$). Substantial agreement was observed for Support Devices ($\kappa = 0.78$), Edema ($\kappa = 0.71$), No Finding ($\kappa = 0.71$), Fracture ($\kappa = 0.70$), Atelectasis ($\kappa = 0.69$), and Consolidation ($\kappa = 0.61$). Lung Lesion ($\kappa = 0.60$) and Lung Opacity ($\kappa = 0.44$) fell in the moderate range. Agreement was weakest for findings defined by composite or inherently subjective criteria: Enlarged Cardiomediastinum ($\kappa = 0.36$), Pneumonia ($\kappa = 0.30$), and Pleural Other ($\kappa = 0.23$), all in the fair-agreement range.

Notably, the low-prevalence findings span the full $\kappa$ spectrum rather than behaving uniformly under chance correction. Fracture, though rare, has a distinctive cortical-disruption appearance and reached substantial agreement ($\kappa = 0.70$), while Lung Lesion achieved moderate agreement ($\kappa = 0.60$). Pleural Other, by contrast, produced only fair agreement ($\kappa = 0.23$) despite its low prevalence, reflecting genuine disagreement about how to label the few positive cases: the category is a composite label encompassing pleural thickening, plaques, calcification, and scarring, whose subtypes readers tend to describe individually rather than under the umbrella term. This pattern indicates that the dominant source of inter-reader disagreement in our corpus is definitional ambiguity of the label itself, not rarity of the underlying finding.

Across the 14 pathologies, Fleiss' $\kappa$ ranged from $0.68$ (PGY-2) to $0.79$ (Attending/Staff), with intermediate values for PGY-3 ($\kappa = 0.73$), PGY-4 ($\kappa = 0.73$), PGY-5 ($\kappa = 0.75$), and Fellow ($\kappa = 0.73$). All cohorts exceeded the substantial-agreement threshold ($\kappa \geq 0.61$). The modest gap between PGY-2 and Attending/Staff suggests a small experience effect on inter-reader consistency, but most of the variability is already resolved by intermediate residency training.

Pairwise agreement between training-level cohorts is shown in Figure~\ref{fig:clinical1} (Panel D). Between-cohort agreement was highest for every pair that included attending/staff radiologists ($\kappa = 0.750$--$0.771$), while resident--resident and resident--fellow pairs were modestly lower ($\kappa = 0.697$--$0.731$). Within-cohort agreement (diagonal) was consistently lower than between-cohort agreement, ranging from $0.503$ (PGY-2) to $0.735$ (Attending/Staff). 

\subsubsection{Agreement between chain-of-thought findings and the original CheXpert Plus report}

We compared pathology mentions extracted from the CoTs against findings documented in the corresponding CheXpert Plus report, characterizing concordance between the independent multi-reader consensus and the original single-reader report (Figure~\ref{fig:clinical1} Panel E). Concordance varied substantially across pathologies. Findings with unambiguous imaging signatures showed the highest sensitivity: Pleural Effusion (82.5\%), Support Devices (79.0\%), Cardiomegaly (74.1\%), and Lung Opacity (70.9\%). Mid-range sensitivity was observed for Edema (64.5\%), Pneumothorax (55.4\%), Atelectasis (50.1\%), Fracture (46.8\%), Consolidation (41.9\%), and No Finding (40.4\%). Findings with ambiguous definitional boundaries or composite categories showed the lowest sensitivity: Pneumonia (37.4\%), Enlarged Cardiomediastinum (35.0\%), Lung Lesion (30.0\%), and Pleural Other (20.2\%). Across training levels, agreement ranged from 61.4\% (Attending/Staff) to 71.2\% (Fellow), with residency cohorts clustered between 65.4\% (PGY-4) and 67.9\% (PGY-3). 

\subsubsection{Agreement on visual attention coordinates}

Spatial agreement between annotators was moderate (mean IoU = 0.60, SD = 0.11, median = 0.60, IQR 0.54--0.68; n = 56,248 pairs). Median IoU was broadly consistent across pathologies, ranging from 0.703 for Lung Lesion (n = 5) to 0.604 for Pneumothorax (n = 342). Among higher-frequency findings, agreement was highest for Edema (0.637), Lung Opacity (0.634), Pleural Effusion (0.624), and No Finding (0.619). Support Devices (n = 50,219) had a median IoU of 0.615, and slightly lower agreement was observed for Pneumonia (0.605), Enlarged Cardiomediastinum (0.605), and Pneumothorax (0.604). Median IoU decreased slightly across training levels (PGY-2: 0.630; PGY-3: 0.628; PGY-4: 0.625; PGY-5: 0.619; fellow: 0.616; attending/staff: 0.602) (Figure~\ref{fig:clinical1} Panel F).

\subsubsection{Semantic similarity of chains-of-thought}

Semantic similarity was computed using ClinicalBERTScore, which quantifies similarity based on contextual embeddings by aligning semantically related tokens between reasoning traces rather than exact word overlap (Figure~\ref{fig:clinical1} Panel G). Across all pairs of independent CoTs generated for the same image, the mean pairwise semantic similarity was modest (mean = 0.350), reflecting substantial variability in how different readers verbalize reasoning about the same study, including differences in ordering, granularity, differential-diagnosis framing, and language style. When CoTs were pooled by annotator continent or training level and similarity was computed between these aggregated group-level reasoning profiles, values were substantially higher, because pooling averages out reader-level stylistic variation and retains shared clinical vocabulary. At the continental level, similarity was highest between North America and Asia (0.945), North America and Oceania (0.932), and Africa and North America (0.924), and lowest between South America and Oceania (0.609) and South America and North America (0.661). Europe and South America demonstrated relatively high similarity (0.876). Similarity by training level was likewise elevated relative to the per-pair mean and was comparable across levels, ranging from 0.359 (attending/staff) to 0.436 (PGY-5).

\subsubsection{Uncertainty expression}

\begin{figure}[H]
\centering
\vspace{-2.3cm}
\makebox[\linewidth][c]{\includegraphics[width=1.96\linewidth]{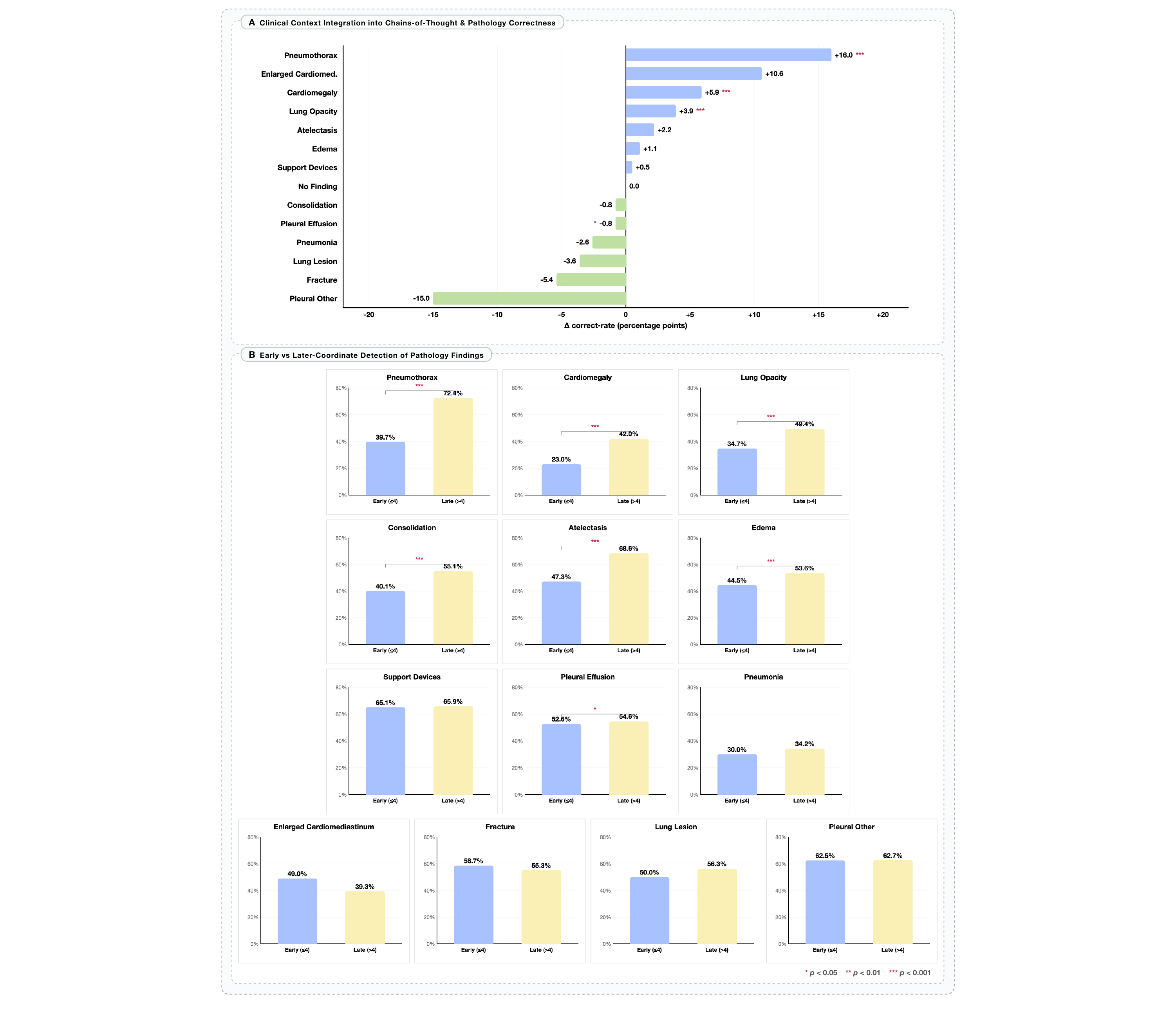}}
\caption{}
\label{fig:clinical1}
\end{figure}

Uncertainty language, operationalized through hedging expressions (e.g., ``may,'' ``likely,'' ``cannot exclude,'' ``suggestive of''), was associated with image quality, diagnostic complexity, and temporal context. Uncertainty was expressed in 44.2\% of CoTs for optimal studies (27,003 of 61,088) versus 84.1\% for suboptimal studies (35,760 of 42,504), and increased with the number of findings: 51.4\% (no findings), 63.6\% (1), 55.3\% (2), 67.0\% (3), 73.2\% (4), and 71.5\% (5+) (Figure~\ref{fig:clinical1} Panel H).

Temporal context amplified uncertainty further (43.3\% of CoTs without temporal references vs.\ 63.8\% with), with the highest rates in CoTs describing worsened cases (76.7\%), followed by improved (67.6\%) and stable (63.7\%). At the pathology level, uncertainty was most frequently expressed for Pneumonia (78.1\%), Consolidation (75.1\%), Lung Opacity (75.0\%), Edema (74.9\%), Fracture (74.5\%), and Pleural Effusion (74.4\%), and least frequently for Pneumothorax (47.7\%) and Enlarged Cardiomediastinum (49.7\%) (Figure~\ref{fig:clinical1} Panel I).

We next examined how verbalized confidence related to agreement between each CoT's findings and the original CheXpert Plus report. Across all CoTs, 54.3\% showed confidence--agreement consistency: either confident CoTs that agreed with the original report, or hedged CoTs that disagreed with it. 23.0\% expressed confidence while disagreeing with the original report, and 22.7\% expressed uncertainty while agreeing with it. Consistency varied minimally by training level (PGY-2: 53.6\%; attending: 54.1\%) but differed substantially by pathology. Consistency was lowest for findings that are typically visually salient when present, Pneumothorax (43.0\%) and Support Devices (42.6\%), where confident CoTs more frequently disagreed with the original reporter. Consistency was highest for findings requiring more integrative interpretation, Pneumonia (61.1\%) and Pleural Thickening (59.5\%), where hedging more often coincided with report agreement.

\begin{figure}[H]
\centering
\vspace{-2.5cm}
\makebox[\linewidth][c]{\includegraphics[width=1.48\linewidth]{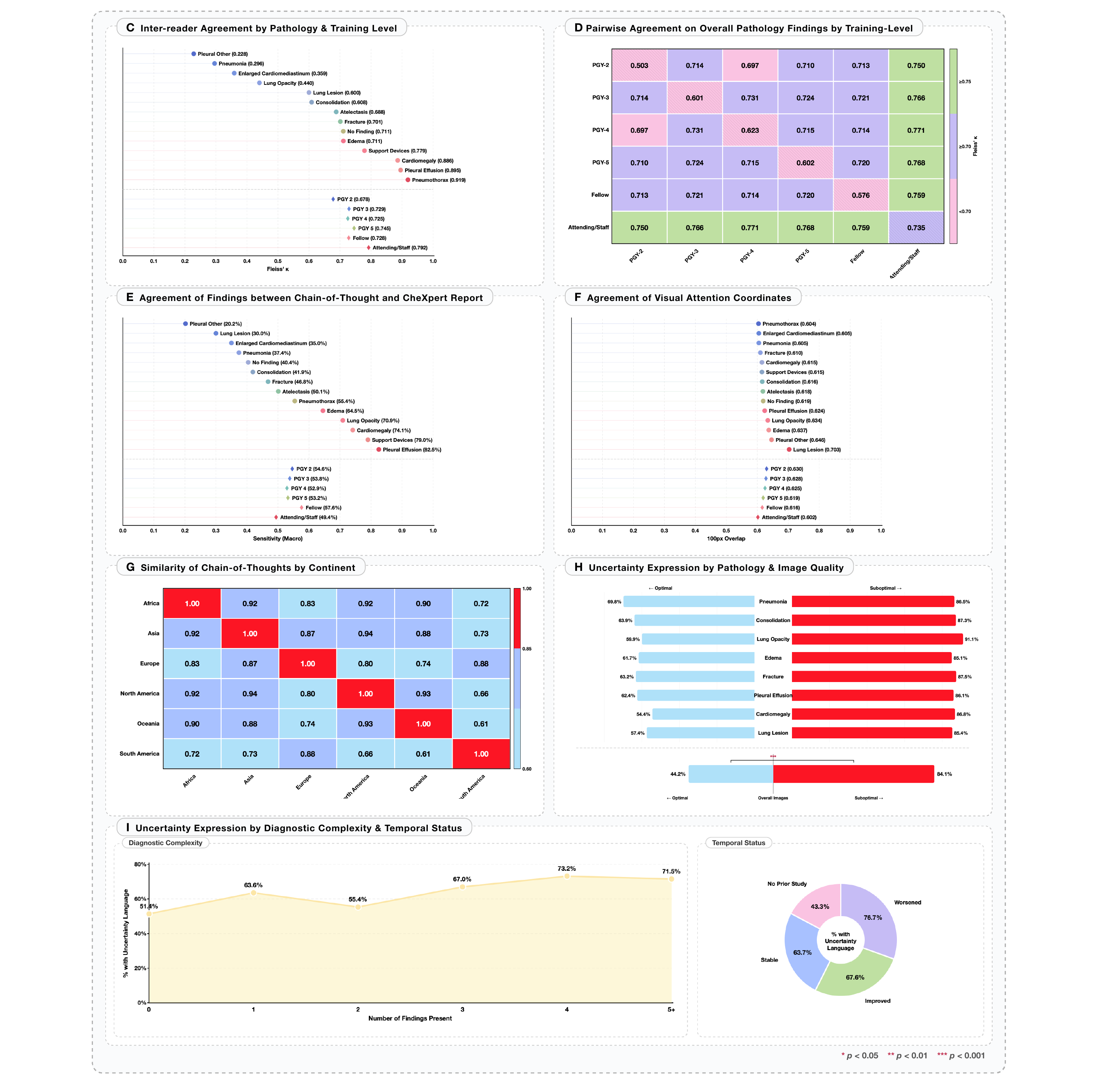}}
\vspace{-1.0cm}
\caption{Clinical Reasoning Patterns in CheXthought \textbf{(A)}~Clinical Context Integration into Chains-of-Thought \& Pathology Correctness. \textbf{(B)}~Early vs Later-Coordinate Detection of Pathology Findings \textbf{(C)}~Inter-reader Agreement by Pathology \& Training Level \textbf{(D)}~Pairwise Agreement on Overall Pathology Findings by Training-Level \textbf{(E)}~Agreement of Findings between Chain-of-Thought and CheXpert Report \textbf{(F)}~Agreement of Visual Attention Coordinates \textbf{(G)}~Similarity of Chain-of-Thoughts by Continent \textbf{(H)}~Uncertainty Expression by Pathology \& Image Quality \textbf{(I)}~Uncertainty Expression by Diagnostic Complexity \& Temporal Status}
\label{fig:clinical2}
\end{figure}

\subsubsection{Universal initiation schema}

To characterize visual attention patterns, we mapped each coordinate to a $4\times 4$ anatomical grid comprising 16 approximate zones spanning apical, upper, mid, and lower lung fields, with right-lateral, right-paramediastinal, left-paramediastinal, and left-lateral columns. Across all 14 pathologies and all six training levels, visual attention was initiated most often in the right upper paramediastinal zone (22--26\% of CoTs) and the left upper paramediastinal zone (16--22\%)---the region encompassing the upper mediastinum, tracheal course, and right hilum. Central zones (both hila, both paracardiac regions, both upper paramediastinal regions) were attended within the first 15--25\% of a CoT's annotations, while peripheral zones (costophrenic angles, upper lateral lung fields) were not reached until 40--48\%. Figure~\ref{fig:visualsearch} (Panel A) illustrates the universal initiation schema, with attention most frequently beginning in the upper paramediastinal and hilar regions. 

\subsubsection{Visual attention clusters into three distinct search-trajectories}
\label{sec:clusters}
\begin{figure}[H]
\vspace{-2.3cm}
\centering
\makebox[\linewidth][c]{\includegraphics[width=1.3\linewidth]{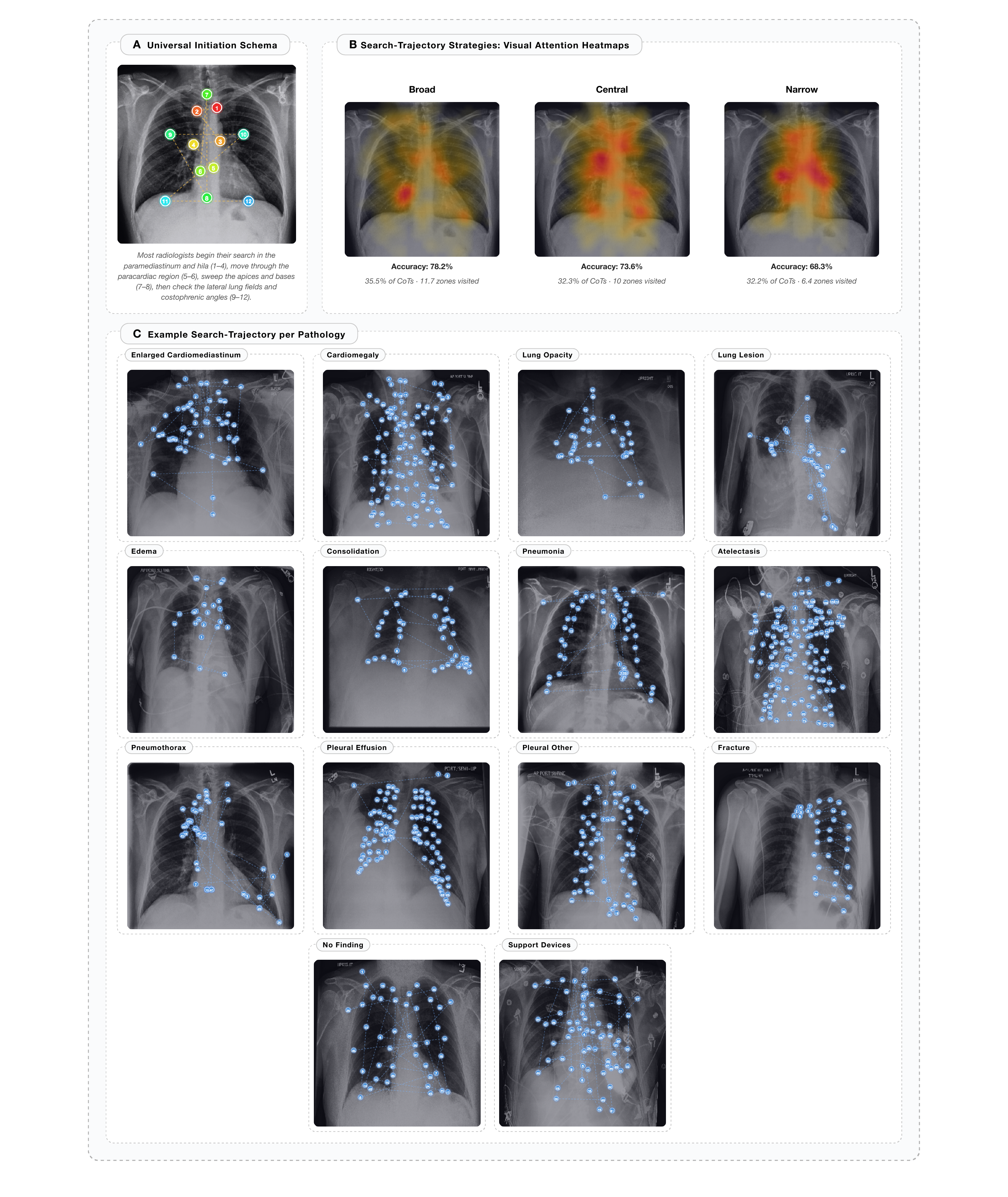}}
\vspace{-1.0cm}
\caption{Visual search strategies derived from spatial attention trajectories \textbf{(A)}~Universal initiation schema of search-trajectory \textbf{(B)}~Three search-trajectory strategies \textbf{(C)}~Representative individual search-trajectories for each of the 14 CheXpert pathology categories}
\label{fig:visualsearch}
\end{figure}

We applied $K$-means clustering to trajectory feature vectors that encode path length, directional entropy, backtracking rate, vertical bias, left-right balance, coverage, revisit rate, and endpoint displacement. Based on the resulting clusters, we labeled each one according to its dominant geometric behavior, shown in Figure~\ref{fig:visualsearch} (Panel B) :
\begin{itemize}
    \item \textbf{Broad (35.5\% of CoTs):} visual attention on the widest range of anatomical zones (mean 11.7 of 16 zones) and extended their attention furthest toward the lung periphery.
    \item \textbf{Central (32.3\%):} focused visual attention on the central mediastinal column and vertically between apical and basal regions (mean 10.0 zones visited), with less exploration of the lateral lung fields.
    \item \textbf{Narrow (32.2\%):} visual attention with the lowest coverage (mean 6.4 zones), remaining in a central region without extending to the periphery.
\end{itemize}
Figure~\ref{fig:visualsearch} (Panel C) provides representative trajectories across each of the 14 pathologies. 
The three strategies differed significantly in diagnostic accuracy, measured as concordance between CoT-mentioned findings and the CheXpert Plus report ($\chi^2(2) = 152.2$, $p < 0.001$, Cram\'er's $V = 0.09$). Broad searchers achieved 78.2\% concordance, compared with 73.6\% for central and 68.3\% for narrow readers. Against the other two strategies combined, broad searchers were 7.3 percentage points more accurate ($z = 10.5$, $p < 0.001$, $h = 0.17$).

\section{Evaluating vision-language reasoning models using CheXthought}

To evaluate the quality of human-authored CoTs in CheXthought and benchmark them against state-of-the-art VLMs, we conducted a blinded comparison study. We selected 300 CXR cases from distinct patients (150 normal and 150 randomly selected pathology cases) from the CheXthought dataset for which human CoTs had been collected during annotation. We then provided the same images to three VLMs: GPT~5.2, Claude Opus~4.5, and MedGemma~1.5, and elicited CoT outputs using the same structured CoT format detailed in Supplementary Table~\ref{tab:CoT}. 

Three attending-level radiologists who had not participated in the CheXthought dataset annotation then evaluated all CoTs in a blinded fashion, scoring each on four dimensions using a 1--5 Likert scale displayed in Supplementary Table~\ref{tab:rubric}. All identifying information was removed, and CoTs were presented in randomized order with anonymized identifiers reshuffled per case to prevent pattern recognition.

Table~\ref{tab:evalresults} displays the performance of VLM-generated CoTs compared with CheXthought dataset CoTs. CheXthought CoTs had the highest overall score (4.81; 95\% CI 4.79--4.84), followed by GPT~5.2 (4.14; 4.10--4.19), MedGemma~1.5 (4.13; 4.08--4.18), and Claude Opus~4.5 (3.89; 3.84--3.94), and outperformed VLMs across every category (Wilcoxon signed-rank all $p < 0.0001$), with the models otherwise performing similarly well. For comprehensiveness of findings and causal support, CheXthought CoTs scored 4.67 and 4.84, respectively, exceeding MedGemma~1.5 for comprehensiveness of findings (4.17) and GPT~5.2 for causal support (4.37). The largest differences were seen for factuality, where CheXthought CoTs scored 4.88 compared with 4.19 for GPT~5.2, 4.12 for MedGemma~1.5, and 3.83 for Claude Opus~4.5, and for spatial localization, where scores were 4.88 for CheXthought CoTs, 4.34 for MedGemma~1.5, 3.79 for GPT~5.2, and 3.50 for Claude Opus~4.5.
 
Within-model comparisons of normal versus pathology performance are shown in Extended Data Table~\ref{tab:normvspath} for the five most common findings. CheXthought CoTs ranked highest across all pathology subgroups and for normal cases.
MedGemma~1.5 performed substantially better on pathology than on normal cases (overall: 4.37 vs.\ 3.92; $p < 0.001$, with markedly higher scores for image grounding (4.47 vs.\ 3.74, $p < 0.001$) and spatial localization (4.62 vs.\ 4.03, $p < 0.001$). GPT~5.2 and Claude Opus~4.5 showed the opposite pattern for spatial localization, scoring substantially lower on pathology than on normal cases (GPT~5.2: 3.42 vs.\ 4.20, $p < 0.001$; Claude Opus~4.5: 2.96 vs.\ 4.07, $p < 0.001$), indicating specific weakness in localizing pathologic findings. All models except MedGemma~1.5 improved on logical consistency when pathology was present, suggesting that the presence of clear findings structures the reasoning chain. CheXthought CoTs were consistently near-ceiling (4.63--5.00) across both case types and all dimensions. Inter-rater agreement across the three radiologists was high, with Gwet's AC2 ranging from 0.69 to 0.96 and exact-or-within-one-point agreement from 45\% to 98\%.

\begin{table}[H]
\centering
\caption{Chain-of-Thought quality scores and illustrative examples. Scores are means across blinded expert reviews (Likert 1--5); 95\% confidence intervals shown in parentheses.}
\label{tab:evalresults}
\scriptsize
\setlength{\tabcolsep}{5pt}
\renewcommand{\arraystretch}{1.25}

\begin{tabular}{@{}lcccc@{}}
\toprule
\textbf{Dimension} & \textbf{CheXthought Data} & \textbf{GPT~5.2} & \textbf{MedGemma~1.5} & \textbf{Claude Opus~4.5} \\
\midrule
Comprehensiveness of Findings & \textbf{4.67 (4.63--4.71)} & 4.25 (4.20--4.30) & 4.17 (4.12--4.22) & 4.03 (3.98--4.08) \\
Causal Support                & \textbf{4.84 (4.81--4.87)} & 4.37 (4.32--4.43) & 3.99 (3.92--4.06) & 4.26 (4.20--4.32) \\
Factuality                    & \textbf{4.88 (4.85--4.90)} & 4.19 (4.14--4.23) & 4.12 (4.04--4.21) & 3.83 (3.76--3.90) \\
Spatial Localization           & \textbf{4.88 (4.85--4.91)} & 3.79 (3.70--3.88) & 4.34 (4.28--4.40) & 3.50 (3.40--3.60) \\
\midrule
\textbf{Overall}              & \textbf{4.81 (4.79--4.84)} & 4.14 (4.10--4.19) & 4.13 (4.08--4.18) & 3.89 (3.84--3.94) \\
\bottomrule
\end{tabular}

\vspace{2pt}
\begin{flushleft}
\scriptsize
All pairwise comparisons $p < 0.0001$. Largest gaps observed in factuality and spatial localization.
\end{flushleft}

\vspace{6pt}

\renewcommand{\arraystretch}{1.3}
\begin{tabular}{@{}l p{5.5cm} p{5.5cm}@{}}
\toprule
\textbf{Dimension} & \textbf{High score (5)} & \textbf{Low score (1)} \\
\midrule
Comprehensiveness &
``Left lower lobe pneumonia with parapneumonic effusion.'' &
``No pleural effusion.'' \textit{[Misses obvious effusion and atelectasis]} \\

Causal Support &
Sharp costophrenic angles leads to no effusion. \textit{[Supported]} &
``No effusion'' leads to ``Suspected effusion.'' \textit{[Contradicted]} \\

Factuality &
``No devices seen.'' \textit{[Correct; none present]} &
``Central venous catheter present.'' \textit{[Hallucinated]} \\

Spatial Localization &
``Right costophrenic blunting.'' \textit{[Correct side]} &
``Right lower lobe opacity.'' \textit{[Actually left]} \\
\bottomrule
\end{tabular}

\end{table}
\section{Developing vision-language reasoning models using CheXthought}
\subsection{Model variants and training design}

Using both chain-of-thought reasoning and visual attention coordinate data from CheXthought, we fine-tuned CheXthought-VLM, a vision-language reasoning model to generate step-by-step diagnostic reasoning in response to chest radiographs. Additionally, using CheXthought's multi-reader annotation data, we trained CheXthought-VLM to predict human–human and AI–human inter-reader disagreement directly from the image.

We trained three additional models using the same base architecture Qwen3-VL-8B-Thinking, and identical image sets from CheXpert Plus, varying only the training signal:

\begin{enumerate}
    \item \textbf{CheXthought-CoT} --- fine-tuned on human-authored expert chain-of-thought reasoning from CheXthought, without visual attention coordinates. This variant isolates the contribution of visual attention beyond chain-of-thought supervision alone.
    \item \textbf{CheXpert-Report} --- fine-tuned on image--report pairs from CheXpert Plus, representing the conventional paradigm of radiology VLM training.
    \item \textbf{Synthetic-CoT} --- fine-tuned on chain-of-thought reasoning traces generated by GPT~5.2 for the same images.
\end{enumerate}

We investigated whether human-authored reasoning improves over synthetic and report-based supervision, and whether human visual attention data provides measurable training benefit beyond chain-of-thought reasoning data alone. All fine-tuned models were trained under identical conditions (bfloat16 precision, batch size~1 with gradient accumulation of 8~steps, AdamW optimizer, learning rate $2\times10^{-5}$, 3~epochs), ensuring that any observed differences were attributable solely to the supervision format. We additionally compare all fine-tuned variants against the unmodified Qwen3-VL-8B-Thinking base model, which serves as a pre-fine-tuning reference point; differences between the base model and each variant reflect the net effect of fine-tuning under a given supervision format.

\subsection{Evaluating visual faithfulness}

We designed four controlled experiments to test whether models genuinely reason about image content or rely on language-based shortcuts. Each experiment isolates a different dimension of visual grounding: text-prior reliance (no image input), spatial grounding (region occlusion), recognition of degraded visual information (image quality), and temporal reasoning across serial studies. We selected 300 chest radiographs from the CheXpert Plus dataset (not used during model training) for the no-image, occlusion, and image-quality experiments. For region occlusion, the diagnostically relevant anatomical region for each labeled pathology was masked with black pixels prior to model inference. For image quality, all 300 images were degraded with Gaussian blur or downsampled to low resolution and re-evaluated. For temporal reasoning, we identified patients with at least two serial chest radiographs from the held-out set and presented each patient's series under three orderings (correct chronological, reversed chronological, and randomized; n = 100 cases per ordering); this experiment was evaluated only for CheXthought-VLM and MedGemma~1.5, as MedGemma~1.5 is the only comparison model trained on temporally ordered multi-study inputs. Pairwise comparisons of model performance on each visual faithfulness test were assessed using McNemar's test across the six model pairs within each condition. All results are displayed in Table~\ref{tab:faithfulness}.\\

\textbf{No image input}
We prompted each model to ``Interpret this CXR and provide step-by-step diagnostic reasoning'' without providing any image. In this setting, any reported pathology finding reflects a hallucinated output, as no visual evidence was available to support diagnostic claims. All models generated pathology findings under this condition every time (100\%).

\textbf{Region occlusion}
We masked the diagnostically relevant anatomical region with black pixels for each pathology case and measured the persistence rate—the proportion of cases in which the model still reports the finding despite its defining region being absent. CheXthought-VLM achieved the lowest overall persistence rate (4\%), with zero persistence for Atelectasis, Cardiomegaly, Edema, and Pneumothorax. CheXthought-CoT showed 6\% overall persistence. Qwen3-VL-8B-Thinking demonstrated an overall persistence of 19.6\%, showing strong regional grounding for Cardiomegaly (0\%) but maintaining high persistence for Pleural Effusion (75\%) and Pneumothorax (41.6\%). MedGemma~1.5 showed the highest overall persistence (47.4\%), with particularly high rates for Pneumothorax (75\%) and Atelectasis (71.4\%), though it performed perfectly on Pleural Effusion (0\%). All comparisons were significant ($p < 0.001$) except CheXthought-VLM (4\%) vs.\ CheXthought-CoT (6\%, $p = 0.46$).

\textbf{Image quality}
To mimic real-world situations where radiologists routinely encounter suboptimal imaging and must recognize when quality limits interpretation, we evaluated models under Gaussian blur and low-resolution degradation, and assessed whether each model recognized that image quality impacted its clinical reasoning. CheXthought-VLM, CheXthought-CoT, and Qwen3-VL-8B-Thinking all identified 100\% of blurred images as suboptimal, while MedGemma~1.5 only recognized 20\%  (all $p < 0.001$). For low-resolution images, CheXthought-VLM showed the highest recognition rate (90\%), followed by Qwen3-VL-8B-Thinking (62\%), CheXthought-CoT (30\%), and MedGemma~1.5 (12\%) ($p < 0.001$).

\textbf{Temporal reasoning}

Comparing current and prior imaging to characterize interval change is central to the radiology workflow. We presented serial studies from the same patient under three orderings: correct chronological, reversed chronological, and randomized order, to test whether models track visual progression rather than assuming a fixed temporal sequence. CheXthought-VLM achieved the highest correct temporal trajectory rate across all conditions: 95\% (95\% CI: 89--98) on correct order, 87\% (95\% CI: 79--92) on reversed, and 80\% (95\% CI: 71--87) on randomized. MedGemma~1.5 showed substantially lower trajectory accuracy: 59\% (95\% CI: 49--68) on correct order, 24\% (95\% CI: 17--33) on reversed, and 43\% (95\% CI: 34--53) on randomized. CheXthought-VLM also referenced temporal comparison in 90\% (95\% CI: 86--93) of responses overall, compared to 42\% (95\% CI: 37--48) for MedGemma~1.5. CheXthought-VLM significantly outperformed MedGemma~1.5 on every condition (McNemar's test, all $p < 0.001$).

\begin{table}[H]
\centering
\caption{Visual Faithfulness Evaluation. Occlusion persistence, image quality recognition, and temporal trajectory accuracy across models. Temporal reasoning evaluated only for CheXthought-VLM and MedGemma~1.5. $^{\dagger}$Significantly outperforms $\geq$3 other models; $^{*}$significantly outperforms $\geq$1 other model (McNemar's test, $p<0.05$). Bold = best per row.}
\label{tab:faithfulness}
\scriptsize
\setlength{\tabcolsep}{4pt}
\renewcommand{\arraystretch}{1.15}
\resizebox{\textwidth}{!}{%
\begin{tabular}{@{}llcccc@{}}
\toprule
\textbf{Test} & \textbf{Condition} & \textbf{CheXthought-VLM} & \textbf{CheXthought-CoT} & \textbf{Qwen3-VL-8B-Thinking} & \textbf{MedGemma~1.5} \\
\midrule
Response rate (\%) & No image & 100 & 100 & 100 & 100 \\
\midrule
False pathology rate & Atelectasis occluded & \textbf{0} & \textbf{0} & 28.2 & 71.4 \\
after occlusion (\%) & Cardiomegaly occluded & \textbf{0} & 20 & \textbf{0} & 14.3 \\
 & Consolidation occluded & 10 & \textbf{0} & 5.0 & 57.1 \\
 & Edema occluded & \textbf{0} & \textbf{0} & 5.6 & 14.3 \\
 & Pleural Effusion occluded & 20 & 10 & 75.0 & \textbf{0} \\
 & Pneumonia occluded & \textbf{0} & 10 & 2.6 & 71.4 \\
 & Pneumothorax occluded & \textbf{0} & \textbf{0} & 41.6 & 75.0 \\
\cmidrule{2-6}
 & Overall & \textbf{4}$^{*}$ & 6$^{*}$ & 19.6$^{*}$ & 47.4 \\
\midrule
Image quality & Gaussian blur & \textbf{100 (96--100)}$^{*}$ & \textbf{100 (96--100)}$^{*}$ & \textbf{100 (96--100)}$^{*}$ & 20 (13--29) \\
acknowledgment (\%) & Low resolution & \textbf{90 (82--95)}$^{\dagger}$ & 30 (21--40)$^{*}$ & 62 (52--71)$^{*}$ & 12 (7--20) \\
\midrule
Temporal trajectory & Correct order & \textbf{95 (89--98)}$^{*}$ & --- & --- & 59 (49--68) \\
accuracy (\%) & Randomized order & \textbf{80 (71--87)}$^{*}$ & --- & --- & 43 (34--53) \\
 & Reversed order & \textbf{87 (79--92)}$^{*}$ & --- & --- & 24 (17--33) \\
\cmidrule{2-6}
Mentioned temporal & Overall & \textbf{90 (86--93)}$^{*}$ & --- & --- & 42 (37--48) \\
comparison (\%) & & & & & \\
\bottomrule
\end{tabular}%
}
\end{table}

\subsubsection{Pathology classification and chain-of-thought quality assessment} 
 
We evaluated these models on chest X-ray pathology classification on two independent external chest X-ray datasets: NIH ChestX-ray14~\citep{NIH2017} and MIMIC-CXR~\citep{Johnson2019}. Models were evaluated under identical inference settings and prompted using the same instruction: ``Interpret this chest X-ray and provide step-by-step diagnostic reasoning followed by your final findings.''
 
Each dataset consisted of 300 images (150 normal and 150 randomly selected pathology cases). Pathology findings were extracted from model outputs using regular expression–based label parsing and compared against ground-truth labels. Performance was evaluated using accuracy (\%) with all metrics computed at the dataset level.
 
To assess chain-of-thought quality, three attending radiologists independently scored each model's outputs evaluating causal support, factuality, and spatial localization (Supplementary Table~\ref{tab:rubric}). Classification performance across the NIH and MIMIC datasets is summarized in Table~\ref{tab:classification}, with chain-of-thought quality scores reported in the lower section of that table. Per-pathology performance breakdowns are provided in Extended Data Table~\ref{tab:PathologyExtendedData}.

CheXthought-VLM achieved the highest overall accuracy at 62.7\% (95\% CI: 59--66\%) and F1 of 59.3\% (95\% CI: 55--64\%), with normal case accuracy of 71.0\% and pathology accuracy of 54.3\%. MedGemma~1.5 ranked second overall at 58.5\% accuracy (95\% CI: 55--62\%) and F1 of 58.7\% (95\% CI: 54--63\%), with normal case accuracy of 58.0\% and pathology accuracy of 59.0\%. CheXthought-CoT achieved 52.3\% overall accuracy (95\% CI: 48--56\%) and F1 of 50.7\% (95\% CI: 46--55\%), with normal case accuracy of 55.7\% and pathology accuracy of 49.0\%. CheXpert-Report achieved 50.5\% overall accuracy (95\% CI: 46--55\%) and F1 of 51.1\% (95\% CI: 47--55\%), with normal case accuracy of 49.3\% and pathology accuracy of 51.7\%. Qwen3-VL-8B-Thinking achieved 40.3\% overall accuracy (95\% CI: 36--44\%) and F1 of only 6.8\% (95\% CI: 4--10\%), with the highest normal case accuracy across all models (76.3\%) but the lowest pathology accuracy (4.3\%). Synthetic-CoT performed lowest overall at 25.0\% accuracy (95\% CI: 22--28\%) and F1 of 18.5\% (95\% CI: 14--23\%), with normal case accuracy of 33.0\% and pathology accuracy of only 17.0\%.
 
On the NIH dataset, CheXthought-VLM achieved the highest overall accuracy (66.3\%) and F1 (63.8\%). CheXthought-CoT was comparable at 64.3\% accuracy and F1 of 59.6\%, while Qwen3-VL-8B-Thinking (94.7\%) led in NIH normal case accuracy but had near-zero pathology accuracy (7.3\%, F1 = 13.0\%). On MIMIC, CheXthought-VLM led in overall accuracy (59.0\%) and MedGemma~1.5 led in F1 (61.6\%), with MedGemma~1.5 achieving the highest pathology accuracy on MIMIC (70.0\%). CheXthought-VLM ranked second on MIMIC pathology accuracy at 49.3\%. Qwen3-VL-8B-Thinking showed a degenerate prediction pattern on MIMIC (pathology accuracy = 1.3\%, F1 = 1.9\%), indicating it classified nearly all MIMIC images as normal. Overall, CheXthought-VLM significantly outperformed CheXthought-CoT ($p < 0.001$), Synthetic-CoT ($p < 0.001$), CheXpert-Report ($p < 0.001$), and Qwen3-VL-8B-Thinking ($p < 0.001$). The difference between CheXthought-VLM and MedGemma~1.5 did not reach significance ($p = 0.108$). 
Chain-of-thought quality was evaluated among the four reasoning models (CheXthought-VLM, CheXthought-CoT, Synthetic-CoT, and Qwen3-VL-8B-Thinking); CheXpert-Report was excluded as it does not generate chain-of-thought reasoning. CheXthought-CoT achieved the highest causal support scores on both NIH (4.36; 95\% CI 4.28--4.44) and MIMIC (4.73; 95\% CI 4.68--4.78), as well as the highest NIH factuality score (4.69; 95\% CI 4.64--4.74). CheXthought-VLM led in spatial localization on both NIH (4.33; 95\% CI 4.24--4.42) and MIMIC (4.26; 95\% CI 4.18--4.34) and achieved the highest MIMIC factuality score (4.63; 95\% CI 4.57--4.69). Both CheXthought-VLM and CheXthought-CoT significantly outperformed Synthetic-CoT and Qwen3-VL-8B-Thinking on every metric across both datasets (Wilcoxon signed-rank,  all $p < 0.001$ except CheXthought-CoT vs.\ Qwen3-VL-8B-Thinking on NIH spatial localization, $p = 0.04$), while differences between the two CheXthought variants were small but generally significant, with CheXthought-CoT favored for causal support and factuality on NIH and CheXthought-VLM favored for spatial localization and MIMIC factuality. Inter-rater agreement across the three radiologists was high, with Gwet's AC2 ranging from 0.64 to 0.85 and exact-or-within-one-point agreement from 31\% to 82\%.

\begin{table}[H]
\centering
\caption{Pathology Classification Performance and Chain-of-Thought Quality.
Accuracy (\%) and F1 score (\%) with 95\% bootstrap confidence intervals on NIH ChestX-ray14 and MIMIC-CXR with each dataset consisting of 300 cases (150 normal and 150 randomly selected pathology cases).
$^{\dagger}$Significantly outperforms $\geq$3 other models; $^{*}$significantly outperforms $\geq$1 other model (McNemar's test, $p<0.05$).
Chain-of-thought quality rated on a 1--5 scale (mean $\pm$ SD). Bold = best per row.}
\label{tab:classification}
\scriptsize
\setlength{\tabcolsep}{3.5pt}
\renewcommand{\arraystretch}{1.15}
\resizebox{\textwidth}{!}{%
\begin{tabular}{@{}lll cccccc@{}}
\toprule
 & \textbf{Dataset} & \textbf{Metric} & \textbf{CheXthought-VLM} & \textbf{CheXthought-CoT} & \textbf{Synthetic-CoT} & \textbf{Qwen3-VL-8B-Thinking} & \textbf{CheXpert-Report} & \textbf{MedGemma~1.5} \\
\midrule
\multirow{12}{*}{\rotatebox{90}{\textbf{Pathology Classification}}}
 & Overall & Accuracy  & \textbf{62.7 (59--66)}$^{\dagger}$ & 52.3 (48--56)$^{*}$ & 25.0 (22--28) & 40.3 (36--44)$^{*}$ & 50.5 (46--55)$^{*}$ & 58.5 (55--62)$^{\dagger}$ \\
 & Overall & F1        & \textbf{59.3 (55--64)} & 50.7 (46--55) & 18.5 (14--23) & 6.8 (4--10) & 51.1 (47--55) & 58.7 (54--63) \\
 & Overall & Normal    & 71.0 (66--76) & 55.7 (50--61) & 33.0 (28--38) & \textbf{76.3 (71--81)} & 49.3 (44--55) & 58.0 (52--64) \\
 & Overall & Pathology & 54.3 (49--60) & 49.0 (43--55) & 17.0 (13--21) & 4.3 (2--7) & 51.7 (46--57) & \textbf{59.0 (53--65)} \\
\cmidrule{2-9}
 & NIH & Accuracy     & \textbf{66.3 (61--72)}$^{\dagger}$ & 64.3 (59--70)$^{\dagger}$ & 28.3 (23--34) & 51.0 (45--57)$^{*}$ & 57.7 (52--63)$^{*}$ & 60.7 (55--66)$^{*}$ \\
 & NIH & F1           & \textbf{63.8 (57--70)} & 59.6 (52--66) & 17.0 (11--23) & 13.0 (6--20) & 54.5 (48--61) & 55.0 (48--62) \\
 & NIH & Normal       & 73.3 (66--80) & 76.0 (69--83) & 42.0 (34--50) & \textbf{94.7 (91--98)} & 64.7 (57--72) & 73.3 (66--80) \\
 & NIH & Pathology    & \textbf{59.3 (51--67)} & 52.7 (45--61) & 14.7 (9--21) & 7.3 (3--12) & 50.7 (43--59) & 48.0 (40--56) \\
\cmidrule{2-9}
 & MIMIC & Accuracy   & \textbf{59.0 (53--65)}$^{\dagger}$ & 40.3 (35--46)$^{*}$ & 21.7 (17--26) & 29.7 (25--35)$^{*}$ & 43.3 (38--49)$^{*}$ & 56.3 (51--62)$^{\dagger}$ \\
 & MIMIC & F1         & 54.6 (48--61) & 43.2 (37--49) & 19.8 (14--26) & 1.9 (0--5) & 48.2 (42--54) & \textbf{61.6 (57--66)} \\
 & MIMIC & Normal     & 68.7 (61--76) & 35.3 (28--43) & 24.0 (17--31) & 58.0 (50--66) & 34.0 (27--42) & \textbf{42.7 (35--51)} \\
 & MIMIC & Pathology  & 49.3 (41--57) & 45.3 (37--53) & 19.3 (13--26) & 1.3 (0--4) & 52.7 (45--61) & \textbf{70.0 (62--77)} \\
\midrule
\multirow{6}{*}{\rotatebox{90}{\textbf{CoT Quality}}}
 & NIH & Causal Support     & 4.08\,$\pm$\,0.90 & \textbf{4.36\,$\pm$\,0.71} & 3.78\,$\pm$\,1.14 & 3.83\,$\pm$\,0.69 & --- & --- \\
 & NIH & Spatial Local.     & \textbf{4.33\,$\pm$\,0.78} & 4.00\,$\pm$\,0.71 & 3.43\,$\pm$\,0.90 & 3.89\,$\pm$\,0.62 & --- & --- \\
 & NIH & Factuality         & 4.50\,$\pm$\,1.02 & \textbf{4.69\,$\pm$\,0.46} & 3.05\,$\pm$\,0.59 & 3.53\,$\pm$\,0.96 & --- & --- \\
\cmidrule{2-9}
 & MIMIC & Causal Support   & 4.68\,$\pm$\,0.57 & \textbf{4.73\,$\pm$\,0.44} & 3.38\,$\pm$\,0.88 & 4.29\,$\pm$\,0.66 & --- & --- \\
 & MIMIC & Spatial Local.   & \textbf{4.26\,$\pm$\,0.72} & 3.84\,$\pm$\,0.75 & 3.14\,$\pm$\,1.10 & 4.17\,$\pm$\,0.37 & --- & --- \\
 & MIMIC & Factuality       & \textbf{4.63\,$\pm$\,0.49} & 4.36\,$\pm$\,0.48 & 3.65\,$\pm$\,0.49 & 3.00\,$\pm$\,1.00 & --- & --- \\
\bottomrule
\end{tabular}%
}
\end{table}

\subsection{Visual attention as a training and inference signal} 

To understand how human visual attention influences model behavior beyond language-level reasoning supervision, we designed a series of experiments to isolate the contribution of visual attention at both training and inference time.

\subsubsection{Training with visual attention supervision does not consistently shift internal visual attention toward human patterns}

\begin{figure}[H]
\centering
\makebox[\linewidth][c]{\includegraphics[width=1.4\linewidth]{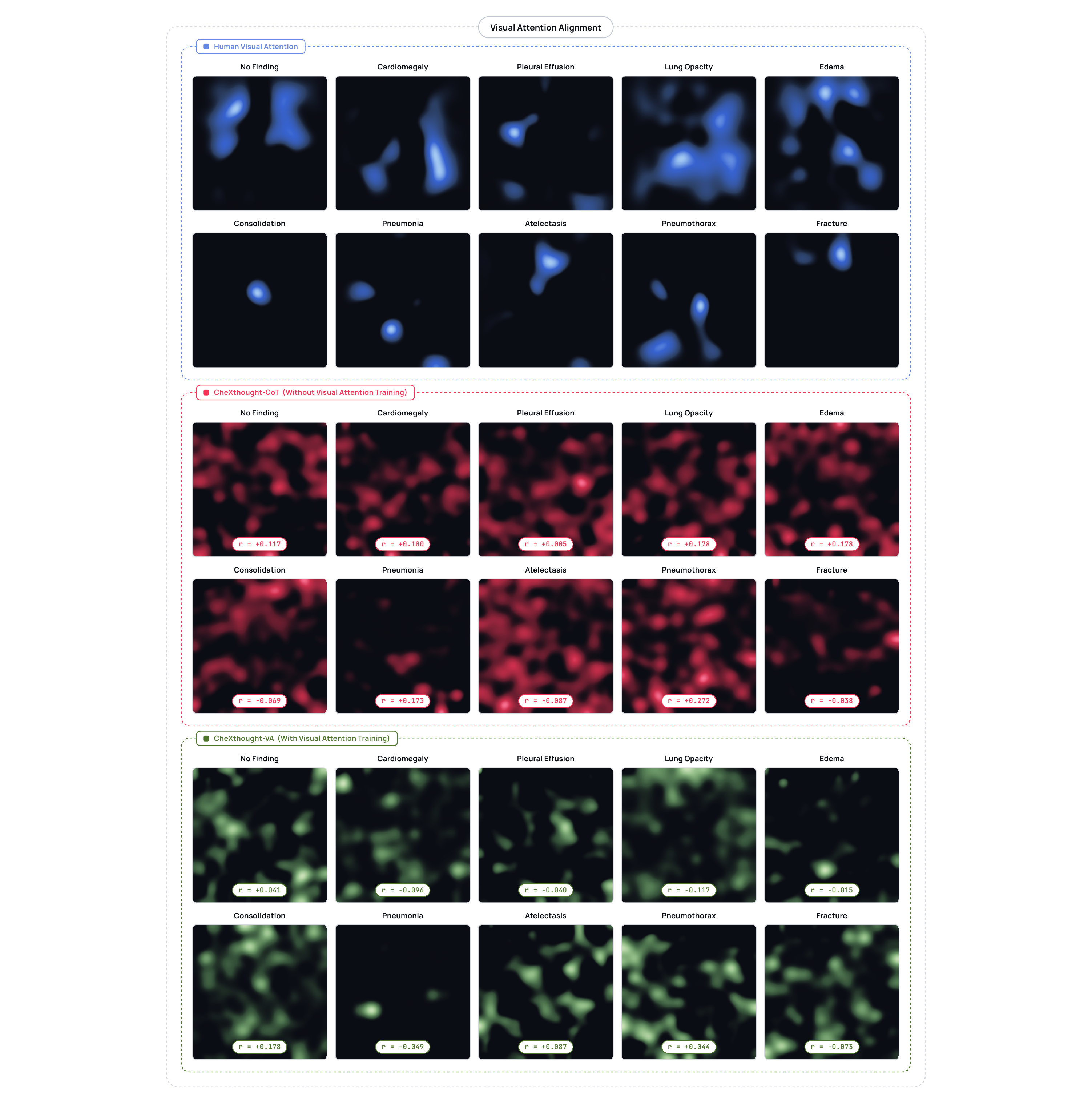}}
\caption{Comparison of human visual attention and Grad-CAM maps for CheXthought models. Pearson correlation ($r$) is reported per pathology.}
\label{fig:gradcam}
\end{figure}

We evaluated whether incorporating radiologist visual attention data in the form of spatial coordinates during training alters a model's internal visual processing. We compared CheXthought-CoT (without visual attention training) to CheXthought-VLM (with visual attention training) by extracting Grad-CAM saliency maps from each model during interpretation of 300 held-out chest X-ray images not seen during training. These saliency maps were compared to corresponding human visual attention from annotators using residual Pearson correlation (Figure~\ref{fig:gradcam}). While alignment improved for some pathologies, including Consolidation and Atelectasis, it decreased for pathologies where the baseline model already showed modest positive alignment, including Lung Opacity, Pneumothorax, Pneumonia, Cardiomegaly, and Edema. Overall, mean residual correlation across all ten pathologies decreased from $r = 0.083$ (CoT-only) to $r = -0.004$ (CoT + Visual Attention). Despite using visual attention supervision during training, the model's internal attention distributions did not reliably move toward those of human radiologists, and in most cases diverged further from the human reference.

\subsubsection{Using search-trajectory as an inference-time hint recovers missed findings and reduces hallucinations}

We investigated whether visual attention can improve model performance at inference time, particularly on cases where models initially fail. We designed inference-time hinting strategies that encoded search-trajectories as structured text, without modifying the input image or retraining the model. We compared the three search-trajectories drawn from the clustering analysis in Section~\ref{sec:clusters}: Broad, Central, and Narrow. As a control, we additionally evaluated a random-dot condition in which spatial coordinates were sampled uniformly at random across the image, instead of following the expert search-trajectories observed in CheXthought.

We evaluated Qwen3-VL-8B-Thinking and MedGemma~1.5 under four experimental conditions (Table~\ref{tab:compact_gaze}) on 300 images (150 normal and 150 randomly selected pathology cases) with each image evaluated under all four strategy conditions and results stratified by baseline performance: (1) cases where the model initially identified no ground-truth pathology (Qwen3-VL-8B-Thinking $n{=}124$, MedGemma~1.5 $n{=}88$), measuring \textit{recovery rate} (proportion of previously missed pathologies now detected) and \textit{hallucination change} ($\Delta$ in hallucinated findings per image); (2) normal cases where the model initially hallucinated findings (no ground-truth pathology to recover) (Qwen3-VL-8B-Thinking $n{=}52$, MedGemma~1.5 $n{=}72$); (3) partial-success cases where the model identified at least one correct finding at baseline (Qwen3-VL-8B-Thinking $n{=}26$, MedGemma~1.5 $n{=}62$), measuring whether guidance further improved detection; and (4) normal cases where the model was already correct at baseline (Qwen3-VL-8B-Thinking $n{=}98$, MedGemma~1.5 $n{=}78$), measuring whether guidance degraded performance by introducing hallucinations.

Across all conditions and both models, expert-derived search-trajectories outperformed the random-dot control. On pathology cases with zero correct findings at baseline, the Narrow strategy achieved 70\% recovery with Qwen3-VL-8B-Thinking and 76\% with MedGemma~1.5, compared to 12\% and 41\% under random dots. On normal cases with no findings, where the model initially hallucinated findings, the Narrow strategy again performed strongest, achieving 83\% and 93\% recovery (i.e., correct reclassification as no findings) with a hallucination reduction of $-2.0$ findings per image for both models, versus 21\% and 14\% under random dots. The Central and Broad strategies both outperformed the random-dot control across most conditions, but underperformed Narrow on targeted recovery, particularly on no finding cases.

On partial-success pathology cases, expert search-trajectories produced modest additional gains of $+0.6$ to $+0.9$ correct findings per image, consistent with the interpretation that the largest benefit of search-trajectory guidance occurs when models fail completely rather than when they have already detected partial pathology. On normal cases where models were already correct at baseline, all search-trajectories introduced small numbers of hallucinated findings ($+0.3$ to $+1.2$ per image), including the random-dot control, suggesting that structured search-trajectory prompting itself carries a modest cost of increased over-reporting on easy cases. Notably, Narrow showed the smallest such cost ($+0.3$ to $+0.5$), indicating that directed search-trajectory guidance preserves model behavior on already-correct cases better than more diffuse search patterns.

Chi-square tests confirmed that recovery rates differed significantly across strategies for both Qwen3-VL-8B-Thinking and MedGemma~1.5 in all Panel~A conditions ($\chi^2 \geq 36.4$, $p < 0.001$, Cram\'{e}r's $V > 0.64$); post-hoc pairwise comparisons with Bonferroni correction showed that Narrow, Central, and Broad each significantly outperformed the random-dot control ($p < 0.001$), with the exception of Broad in two subconditions ($p > 0.05$). In Panel~B, all expert strategies produced significantly greater gains in correct findings than Random on partial-success cases ($p < 0.05$), while hallucination increases on already-correct normal cases were significantly lower for Narrow than Random with Qwen3-VL-8B-Thinking ($p < 0.001$) and significantly higher for Central and Broad than Random with MedGemma~1.5 ($p < 0.001$).

\begin{table}[H]
\centering
\caption{Impact of search-trajectory clusters on diagnostic recovery and hallucination suppression.
Recovery\% = proportion of baseline-failing cases now correctly classified;
$\Delta$Corr = change in correct findings per image;
$\Delta$Hall = change in hallucinated findings per image.
$^{*}p < 0.05$, $^{***}p < 0.001$ vs.\ Random (control), Bonferroni-corrected.
$^{\dagger}$Sig.\ \textit{more} hallucinations than Random.
$^{\ddagger}$Sig.\ \textit{fewer} hallucinations than Random.}
\label{tab:compact_gaze}
\scriptsize
\setlength{\tabcolsep}{3.5pt}
\renewcommand{\arraystretch}{1.15}
\resizebox{\textwidth}{!}{%
\begin{tabular}{@{}ll cccc@{}}
\toprule
& & \multicolumn{2}{c}{\textbf{Qwen3-VL-8B-Thinking}} & \multicolumn{2}{c}{\textbf{MedGemma~1.5}} \\
\cmidrule(lr){3-4} \cmidrule(lr){5-6}
\textbf{Condition} & \textbf{Strategy} & Recovery/\,$\Delta$Corr & $\Delta$Hall & Recovery/\,$\Delta$Corr & $\Delta$Hall \\
\midrule
Pathology, 0 correct  & Central          & 68\%$^{***}$  & $-1.9$ & 78\%$^{***}$  & $-2.0$ \\
                      & Narrow           & 70\%$^{***}$  & $-1.4$ & 76\%$^{***}$  & $-1.2$ \\
                      & Broad            & 67\%$^{***}$  & $-2.0$ & 54\%           & $-2.0$ \\
                      & Random (control) & 12\%          & $-0.7$ & 41\%           & $-0.3$ \\
\midrule
No Finding, 0 correct & Central          & 56\%$^{***}$  & $-1.3$ & 75\%$^{***}$  & $-1.5$ \\
                      & Narrow           & 83\%$^{***}$  & $-2.0$ & 93\%$^{***}$  & $-2.0$ \\
                      & Broad            & 44\%           & $-1.3$ & 58\%$^{***}$  & $-1.3$ \\
                      & Random (control) & 21\%          & $-0.7$ & 14\%           & $-0.3$ \\
\midrule
\midrule
Pathology, $\geq$1 correct & Central          & +0.6$^{*}$    & +0.5   & +0.9$^{***}$  & +1.2$^{\dagger}$ \\
                           & Narrow           & +0.6$^{*}$    & +0.4   & +0.7$^{***}$  & +1.1$^{\dagger}$ \\
                           & Broad            & +0.8$^{***}$  & +0.5   & +0.9$^{***}$  & +1.1$^{\dagger}$ \\
                           & Random (control) & +0.1          & +0.3   & +0.2           & +0.5 \\
\midrule
No Findings, correct            & Central          & ---   & +0.9              & ---   & +1.2$^{\dagger}$ \\
                           & Narrow           & ---   & +0.3$^{\ddagger}$ & ---   & +0.5 \\
                           & Broad            & ---   & +1.0              & ---   & +1.2$^{\dagger}$ \\
                           & Random (control) & ---   & +0.8              & ---   & +0.6 \\
\bottomrule
\end{tabular}%
}
\end{table}
\subsection{Human-human and AI-human disagreement prediction}

Using CheXthought's multi-read cases (2--36 independent readers per image), we explored inter-reader disagreement as a potential proxy for case complexity and diagnostic uncertainty, consistent with prior work suggesting that variability across readers can reflect underlying ambiguity \citep{Zinovev2012, Espeland2013}. For each image, we computed the proportion of readers who disagreed on the presence of each finding, yielding a per-finding human--human disagreement rate. We additionally computed per-finding AI--human and AI--AI disagreement rates using three vision--language models (GPT~5.2, Claude Opus~4.5, and MedGemma~1.5 ), aggregated across all findings per image to produce continuous disagreement scores. CheXthought-VLM was then fine-tuned to predict both human--human and AI--human disagreement directly from the image, with performance and calibration assessed on a held-out test set of 300 images from CheXthought.

CheXthought-VLM's predicted disagreement rates were closely aligned with observed rates on both tasks (Table~\ref{tab:disagreement}). For the human-human task, the observed mean disagreement rate was 31.8\% compared to a predicted mean of 36.0\%, a deviation of $+4.2$ percentage points. For the AI-human task, the observed mean was 32.3\% compared to a predicted mean of 35.1\%, a deviation of $+2.8$ percentage points.
 
At the pathology level, human-human predictions were closely aligned with observed disagreement for Edema (observed 35.1\% vs.\ predicted 36.9\%; deviation $+1.8$), Consolidation (32.7\% vs.\ 34.7\%; $+2.0$), and Pneumothorax (37.6\% vs.\ 35.2\%; $-2.4$). Larger deviations occurred for No Finding (26.4\% vs.\ 39.2\%; $+12.8$), Pleural Effusion (29.8\% vs.\ 38.3\%; $+8.5$), Cardiomegaly (29.1\% vs.\ 36.7\%; $+7.6$), Atelectasis (30.8\% vs.\ 38.2\%; $+7.4$), and Lung Opacity (32.5\% vs.\ 38.7\%; $+6.2$), all in the direction of over-prediction. Pneumonia was the only pathology under-predicted (32.6\% vs.\ 26.6\%; $-6.0$). AI-human predictions tracked observed rates within $\pm5$ percentage points for six of nine pathologies (Edema, Consolidation, Pneumothorax, Pneumonia, Cardiomegaly, and Pleural Effusion), with slightly larger over-predictions for Lung Opacity ($+7.3$), Atelectasis ($+6.9$), and No Finding ($+6.3$).
 
Comparing the three observed disagreement signals, overall rates were remarkably similar (human-human 31.8\%, AI-human 32.3\%, AI-AI 33.8\%) but diverged substantially at the pathology level, indicating that AI-AI concordance does not uniformly track either AI-human or human-human concordance. The clearest divergences occurred for Atelectasis (HH 30.8\%, AI-H 38.4\%, AI-AI 52.9\%), where AI systems disagreed with one another far more than with human readers; Pneumothorax (HH 37.6\%, AI-H 34.0\%, AI-AI 20.9\%), where AI systems agreed with one another more than with human readers; and No Finding (HH 26.4\%, AI-H 44.1\%, AI-AI 24.4\%), where AI systems agreed with one another at rates close to human-human agreement but diverged from the human consensus. 
 
\begin{table}[H]
\centering
\caption{Predicted vs.\ observed inter-reader disagreement rates by pathology. ``GT'' denotes observed ground truth; ``Pred'' denotes CheXthought-VLM model prediction. Human-Human rates are computed on the 300-image held-out CheXthought test set; AI-Human and AI-AI observed rates are computed on 604 external chest radiographs (MIMIC-CXR + NIH ChestX-ray14) across GPT~5.2, Claude Opus~4.5, and MedGemma~1.5 .}
\label{tab:disagreement}
\scriptsize
\setlength{\tabcolsep}{6pt}
\renewcommand{\arraystretch}{1.15}
\begin{tabular}{@{}lcc|cc|c@{}}
\toprule
 & \multicolumn{2}{c|}{\textbf{Human-Human (\%)}} & \multicolumn{2}{c|}{\textbf{AI-Human (\%)}} & \textbf{AI-AI (\%)} \\
\textbf{Pathology} & \textbf{GT} & \textbf{Pred} & \textbf{Obs.} & \textbf{Pred} & \textbf{Obs.} \\
\midrule
\textit{Overall} & \textbf{31.8} & \textbf{36.0} & \textbf{32.3} & \textbf{35.1} & \textbf{33.8} \\
\midrule
No Finding        & 26.4 & 39.2 & 44.1 & 50.4 & 24.4 \\
Edema             & 35.1 & 36.9 & 28.5 & 32.4 & 33.1 \\
Consolidation     & 32.7 & 34.7 & 23.8 & 20.3 & 24.5 \\
Pneumothorax      & 37.6 & 35.2 & 34.0 & 36.4 & 20.9 \\
Lung Opacity      & 32.5 & 38.7 & 31.1 & 38.4 & 37.5 \\
Pneumonia         & 32.6 & 26.6 & 20.0 & 23.3 & 28.1 \\
Atelectasis       & 30.8 & 38.2 & 38.4 & 45.3 & 52.9 \\
Cardiomegaly      & 29.1 & 36.7 & 27.9 & 30.2 & 37.4 \\
Pleural Effusion  & 29.8 & 38.3 & 42.5 & 39.2 & 45.8 \\
\bottomrule
\end{tabular}
\end{table}

\section{Discussion}

In this work, we introduce, evaluate and demonstrate the utility of CheXthought, a large-scale human-authored chain-of-thought dataset, grounded through synchronized visual attention, linking verbalized reasoning to pathology and anatomical locations for CXR interpretation. With every image multi-read and interpreted by at least 2 and up to 36 annotators, CheXthought offers insight into the cognitive and visual attention of 501 board-certified radiologists, radiology fellows, and radiology residents across 71 countries, representing the range of expertise involved in chest X-ray interpretation in the real-world clinical setting. Our participatory approach instead incorporates the global radiology workforce as active contributors, embedding geographic and demographic diversity and a broader range of clinical perspectives into model development \citep{Kuhlman2020}. Most existing radiology datasets are annotated by a small number of radiologists from single institutions or non-radiologist personnel \citep{Myronenko2025, Sambara2025, Baharoon2025}. This limited diversity can introduce systematic biases in labeling and interpretation, which may be inherited and amplified by models trained on such data. 

The results of our inter-rater analyses mirror real-world patterns of diagnostic variability and visually grounded reasoning that are well recognized in radiology practice but are not captured in current CXR datasets or multimodal models \citep{CheXpert2019, NIH2017, Johnson2019, Nguyen2022, Nam2025}. Capturing this variability is vital for VLM development, as it exposes models to the inherent ambiguity they will encounter upon deployment \citep{Quinn2023}. This variability manifests in two clinically distinct dimensions that CheXthought is uniquely positioned to address: diagnostic uncertainty at the individual reader level, and systematic patterns of disagreement across readers and between readers and AI systems.

At the level of individual interpretation, we found that uncertainty was significantly more prevalent in CoTs associated with suboptimal image quality and difficult pathologies, including pneumonia. In clinical settings, diagnostic information is frequently compromised by technical factors such as patient positioning or motion artifacts \citep{Kjelle2022}. While we demonstrate that standard vision--language models often fail to recognize when an image is technically inadequate, CheXthought-VLM, trained on CheXthought CoT and visual attention data which explicitly verbalize image quality and technical adequacy, enables the model to acknowledge diagnostic limitations. Despite its importance, uncertainty is still underexplored in existing CXR datasets, with the CheXpert dataset being among the few large-scale CXR datasets that explicitly include uncertainty labels as part of its labeling schema. However, this label captures only the output of an automated rule-based labeler applied to report text, not the underlying cognitive basis for that uncertainty and what visual features made the finding difficult to determine \citep{CheXpert2019}. This distinction is especially important in the era of generative AI. Unlike traditional discriminative models, generative models can produce fluent, persuasive, and highly authoritative outputs even when their reasoning is flawed or their conclusions are wrong, making confident inaccuracy dangerous in healthcare, where models encountering unfamiliar populations or ambiguous cases continue to make confident predictions despite operating outside their training distribution \citep{Ji2023, Celi2025}. The ability of AI systems to communicate uncertainty is a clinical necessity, as interpretive error in radiology remains a persistent challenge, with perceptual errors accounting for 60–80\% of diagnostic mistakes and overall error rates remaining largely unchanged despite dramatic advances in imaging technology \citep{Waite2017, Berlin2007}. These errors result in delayed diagnosis, inappropriate management, and patient harm, while imposing a substantial financial burden on the healthcare system, with the annual cost of measurable medical errors estimated at \$17.1 billion in the United States alone \citep{VanDenBos2011}. AI models that fail to express uncertainty risk compounding rather than mitigating this burden, presenting incorrect predictions with the same authority as correct ones and potentially reinforcing the very diagnostic errors they are intended to reduce. CheXthought addresses this by training on chain-of-thought reasoning that explicitly verbalizes diagnostic uncertainty and its visual basis, enabling CheXthought-VLM to learn not only what is uncertain but why, grounding uncertainty in specific image features and clinical considerations rather than reducing it to a scalar confidence score.

At the level of inter-reader and AI–human comparison, we train CheXthought-VLM to predict both human–human and AI–human disagreement directly from the image, enabling the model to distinguish between cases that are inherently ambiguous to expert readers and cases where the model itself is likely to deviate from expert consensus. These two qualitatively distinct failure modes are not captured by standard scalar confidence estimates, which have been shown to inadequately represent predictive uncertainty in medical machine learning \citep{Kompa2021}. Although aggregate disagreement rates were broadly similar across human–human, AI–human, and AI–AI comparisons, divergence at the pathology level was substantial, with certain findings showing AI systems disagreeing with one another far more than human readers did, and others showing AI systems reaching internal consensus while systematically diverging from the human consensus, a pattern that would not be detectable through AI–AI consistency checks alone.

The need for transparent disagreement modeling is especially urgent given the accelerating pace at which AI tools are being deployed across healthcare systems without a consistent framework for post-deployment surveillance, benchmarking, or evidence generation to support claims of clinical value \citep{NatureMed2026}. Rather than treating every model output as equally authoritative, disagreement prediction explicitly communicates both the difficulty of a case and the trustworthiness of the model's prediction, enabling dynamic control over how outputs are surfaced, prioritized, or withheld in clinical workflows. High predicted human–human disagreement may identify cases that warrant more experienced review, while high predicted AI–human disagreement flags cases where the model's output should not be used as a primary decision support signal. Prior work has shown that integrating disagreement prediction into AI-assisted pipelines can guide clinician attention, trigger second reads, and reduce both automation bias and clinician burden by selectively elevating uncertain or high-risk findings \citep{Sanchez2023}. Our approach enables stratification of clinical queues by both case difficulty and model reliability, aligning model behavior with real-world diagnostic variability and making uncertainty more accurate, interpretable, and actionable for clinicians \citep{Goddard2012, Lyell2017, Yu2025}.

In our experimental design, we sought to isolate and quantify the impact of chain-of-thought reasoning and visual attention as supervision signals on model performance across multiple tasks. By holding the model architecture and training data constant and varying only the form of supervision, we were able to directly assess the contribution of each component. We find that chain-of-thought supervision improves pathology classification relative to report-only training, and that combining chain-of-thought with visual attention leads to further gains, suggesting that these signals provide complementary information rather than redundant structure.

Notably, these improvements in the model occur without alignment in visual attention. Grad-CAM analysis shows that even after training with large-scale visual attention data, model attention maps do not meaningfully align with expert visual search patterns and do not shift following visual attention supervision. This indicates that visual attention supervision influences model behavior without inducing human-like visual search.

Across the visual faithfulness experiments, all models remain susceptible to hallucination when image information is absent. However, CheXthought-VLM, trained with human-authored reasoning and visual attention data, exhibited relatively fewer hallucinations and greater sensitivity to the presence or absence of visual evidence, suggesting a stronger coupling between its predictions and localized image features. This effect is particularly pronounced in the occlusion experiments, where models trained without chain-of-thought reasoning and visual attention are more likely to exhibit unfaithfulness and persist in reporting findings when the diagnostically relevant region is removed. 

CheXthought captures temporal reasoning over serial images, reflecting real-world clinical practice where radiologists rarely interpret chest X-rays in isolation, but instead compare current and prior studies to assess interval change and refine diagnostic confidence \citep{White1994}. This type of longitudinal reasoning is largely absent from existing medical imaging datasets, creating a ``missing context'' problem \citep{StanfordHAI2025}. When presented with serial studies in reversed or randomized order, CheXthought-VLM demonstrated a robust ability to track progression of findings based on visual features rather than assuming a standard chronological flow. This suggests that the model is utilizing visual grounding to anchor its comparisons in the actual image and not just relying on heuristics. 

Several prior datasets have attempted to capture aspects of visual reasoning in medical imaging, particularly within the medical visual question answering (Med-VQA) format \citep{Lau2018}. Early datasets such as VQA-RAD and later large-scale resources including PMC-VQA and OmniMedVQA introduced question--answer supervision for evaluating visual reasoning capabilities \citep{Zhang2023, Hu2024}. More recent work has incorporated structured chain-of-thought rationales and region-grounded reasoning sequences \citep{Sambara2025}. However, these datasets frame reasoning within the context of predefined questions rather than authentic diagnostic interpretation. Similarly, visual grounding in existing medical datasets has largely taken one of two forms: eye-gaze trajectories captured through eye-tracking studies or static bounding box annotations around pathological findings. The REFLACX dataset recorded synchronized eye-tracking data from five radiologists interpreting 2,616 chest X-rays \citep{Lanfredi2022}, while the Eye Gaze Data for Chest X-rays dataset captured eye-gaze trajectories from a single radiologist interpreting 1,083 studies \citep{Karargyris2021}. Although these datasets demonstrated that eye-gaze patterns can approximate regions of visual attention, they primarily reflect unconscious scanning behavior rather than deliberate spatial references tied to diagnostic reasoning. Conversely, datasets such as VinDr-CXR and MS-CXR provide bounding box annotations identifying pathological findings \citep{Nguyen2022}, but these annotations are static localization labels that are not linked to the reasoning processes used to interpret the image.

There are several limitations in our dataset that warrant acknowledgment. First, our cursor-based spatial annotation approach provides an approximation of visual attention rather than true eye-gaze data. While cursor position captures spatial visual attention, it may not fully reflect the rapid, often unconscious saccadic patterns that characterize expert image search found in eye-gaze. Second, both our chain-of-thought quality and fine-tuning evaluations focused on specific VLMs at a single point in time. Given the rapid pace of model development and improvement, performance rankings may shift with newer model versions. Third, our structured CoT annotation framework may have impacted the natural variability of expert reasoning. 

Overall, CheXthought captures dimensions of diagnostic reasoning that are largely absent from existing CXR benchmarks, including inter-reader variability, uncertainty, temporal trajectory assessment of patients with serial images, and spatially grounded rationales linking verbalized reasoning to anatomical and pathology locations. Our findings suggest that authentic, visually grounded human reasoning is a valuable and currently underutilized training signal for building more reliable and interpretable medical vision-language models.

\section{Methods}

All annotators provided informed consent before participating in this study, approved by the Institutional Review Board of Stanford University (IRB approval number 82421). Our infrastructure supported multilingual collaboration, with radiologists recording their interpretations in English, Turkish, German, French, Spanish, Russian, Korean, Dutch, Bulgarian, and Portuguese. Although all annotators possessed proficient English skills, they were encouraged to give their interpretation of cases in English and then in their native language. Non-English CoTs that were not accompanied by an English translation were translated using Gemini 3 Pro, and subsequently verified by a native speaker of the source language to ensure accuracy of medical terminology and reasoning content. To mirror clinical practice, in which radiologists routinely form an independent impression before consulting a colleague, a resource, or prior reports, the original CheXpert Plus radiology report was made accessible in a side panel after approximately 2 minutes of independent interpretation. For serial imaging cases assigned in chronological order, annotators had already viewed reports from earlier studies in the sequence and could reference them when interpreting subsequent images. Additionally, the platform was structured with dedicated input fields corresponding to these components, allowing annotators to explicitly document each element of their reasoning. While there is overlap between aspects of the CoT and the radiology reports, such as findings and temporal comparisons, we specifically instructed annotators and designed our CoT framework to detail the reasoning behind those conclusions, capturing what is not included in radiology reports. Cases were randomly assigned to annotators, with the constraint that serial images from a given patient were assigned to the same annotator and presented in chronological order, with a mean of 2.51 studies per patient. Annotators could not self-select cases.
To ensure data quality, the platform automatically flagged and rejected non-informative or incomplete submissions (e.g., empty responses, placeholder text, or entries lacking sufficient reasoning or corresponding visual tracing). Importantly, these checks did not impose constraints on diagnostic content, reasoning structure, or length beyond a minimal threshold, thereby preserving authentic inter-reader variability. Visual attention coordinates were recorded in pixel space and normalized to the range [0, 1] relative to image width and height, ensuring consistent spatial analysis across images of varying resolution. All outputs underwent review by a panel of 20 attending, board-certified radiologists, with cases distributed among reviewers, to ensure fidelity to the original reasoning. 

\subsection{Statistical analysis} 
 
Analyses were conducted at the level of individual CoT interpretations unless otherwise specified. Descriptive statistics were used to summarize dataset characteristics, including the number of CoTs, spatial annotations, unique images, and participating radiologists. Continuous variables, including CoT word length and spatial annotation counts, were summarized using medians and interquartile ranges (IQRs) due to non-normal distributions. Comparisons of CoT length and spatial annotation density were stratified by annotator training level and pathology category.
 
Inter-rater agreement for pathology identification within CoT interpretations was quantified using Fleiss' $\kappa$ across images interpreted by two or more annotators. Binary labels were generated to indicate whether each pathology was explicitly identified as present in the CoT narrative. Agreement estimates were reported overall and stratified by pathology, annotator training level, and training-level pair combinations. Ninety-five percent confidence intervals were estimated using nonparametric bootstrap resampling of images (1,000 iterations). Agreement on findings was also computed, defined as the probability that two annotators independently labeled the same pathology as definitively present for a given image, summarized by pathology category and training level. Spatial agreement between annotators was quantified using Intersection over Union (IoU) on binarized coordinate heatmaps, reported by pathology and training level. Semantic similarity between CoTs was evaluated using Bio\_ClinicalBERT-based cosine similarity.
 
To evaluate concordance between CoT-derived findings and radiology reports, pathology mentions extracted from each CoT narrative were compared with labels derived from the corresponding CheXpert Plus report. Concordance was defined as the proportion of CoT interpretations that correctly identified a pathology labeled as present in the reference report. Concordance estimates were summarized by pathology and annotator training level, with bootstrap-derived confidence intervals.
 
For CoT quality evaluation, three attending radiologists independently scored blinded, randomized CoT outputs on four dimensions (comprehensiveness of findings, causal support, factuality, spatial localization) using a 1--5 Likert scale. Scores were summarized as means with pairwise comparisons between sources. For pathology classification, accuracy was computed overall and stratified by dataset, case type, and individual pathology, with 95\% confidence intervals estimated via bootstrap resampling (10,000 iterations). Concordance between model saliency (Grad-CAM) and human visual attention was assessed using Pearson correlation, Normalized Scanpath Saliency, KL divergence, and top-20\% spatial overlap. For visual faithfulness experiments, persistence rate, image quality recognition rate, and temporal trajectory accuracy were computed as proportions with bootstrap confidence intervals where applicable. For disagreement prediction, calibration was assessed by computing the deviation between predicted and observed inter-reader disagreement rates overall and by pathology. Analyses were conducted in Python (version 3.10) using NumPy, pandas, and SciPy.

\section{Data availability}

103,592 chains-of-thought comprising the CheXthought dataset are publicly available through the Stanford Center for Artificial Intelligence in Medicine and Imaging, with an additional 10\% sequestered as a held-out test set to support future benchmarking.

\section{Code availability}

CheXthought-VLM is available at https://huggingface.co/StanfordAIMI/CheXthought

\clearpage
\bibliographystyle{plainnat}

\clearpage

\section{Ethics declaration} 

\paragraph{} \textbf{C.P.L.} has the following personal financial interests that are not related to this article: has received research funding to his institution from AWS; Bunkerhill Health; Carestream; CARPL.ai; Clairity; GE Healthcare; Google Cloud; IBM; Kheiron; Lambda; Lunit; Microsoft; Nightingale Open Science; Philips; Siemens Healthineers; Stability.ai; Subtle Medical; VinBrain; Visiana; Whiterabbit.ai; the National Institute of Biomedical Imaging and Bioengineering; the National Heart, Lung, and Blood Institute; the Advanced Research Projects Agency for Health; and the Agency for Healthcare Research and Quality. He reports consultant fees from Sixth Street and Gilmartin Capital; speaking fees and honoraria from the Singapore Ministry of Health, Philips Medical, Canon Medical, and McKinsey \& Company outside the submitted work; and equity in Bunkerhill Health, Sirona Medical, Whiterabbit.ai, Galileo CDS, ADRA.ai, RadPartners, Abxtract, Harrison.ai, and TurboRadiology.
\paragraph{} \textbf{G.S.} is a Shareholder and Consultant for MD.ai, on the SIIM Board of Directors, and is a MIDRC investigator, funded by the National Institutes of Health (NIH) and ARPA-H. 
\paragraph{}The other authors declare no competing interests.

\section{Extended data}

\setcounter{figure}{0}
\setcounter{table}{0}
\renewcommand{\figurename}{Extended Data Figure}
\renewcommand{\tablename}{Extended Data Table}

\begin{table}[H]
\centering
\caption{Mapping of raw findings to CheXpert labels and unmapped findings. $N_{\text{mentions}}$ is the summed count across all synonymous raw findings.}
\label{tab:ExtendedData1}
\scriptsize
\setlength{\tabcolsep}{4pt}
\begin{tabular}{@{}llr@{}}
\toprule
\textbf{CheXpert Label} & \textbf{Raw Findings Mapped} & \textbf{$N_{\text{mentions}}$} \\
\midrule
No Finding & No Finding & 37,637 \\
Cardiomegaly & Cardiomegaly & 34,943 \\
Pleural Effusion & Pleural Effusion & 22,347 \\
Edema & \parbox[t]{7.5cm}{Pulmonary Edema, Pulm.\ Vascular Congestion, Cephalization} & 4,783 \\[6pt]
Enlarged Cardiomediastinum & Widened Mediastinum, Mediastinal Abnormality & 4,488 \\
Lung Opacity & \parbox[t]{7.5cm}{Lung Opacity, Infiltrate, Reticular Pattern, Interstitial Lung Disease, Pulmonary Fibrosis} & 3,987 \\[6pt]
Support Devices & \parbox[t]{7.5cm}{Central Venous Catheter, NG/OG Tube, Pacemaker/ICD, Endotracheal Tube, Chest Tube, Tracheostomy, Orthopedic Hardware, Drain, Stent, PICC Line, CABG, Spinal Hardware, Port, Prosthesis, Sternotomy} & 3,562 \\[6pt]
Atelectasis & Atelectasis, Volume Loss & 3,127 \\
Consolidation & Consolidation & 2,419 \\
Pneumothorax & Pneumothorax & 1,823 \\
Pneumonia & Pneumonia & 1,258 \\
Fracture & \parbox[t]{7.5cm}{Fracture, Rib Fracture, Clavicle Fracture, Vertebral Compression Fx} & 892 \\[6pt]
Lung Lesion & \parbox[t]{7.5cm}{Pulmonary Nodule, Lung Mass, Cavitary Lesion, Granuloma} & 347 \\[6pt]
Pleural Other & Pleural Thickening & 284 \\
\midrule
\multicolumn{2}{@{}l}{\textbf{Subtotal mapped (42 findings)}} & \textbf{121,897 (96.6\%)} \\
\midrule
\multicolumn{3}{@{}l}{\textit{Unmapped findings}} \\[3pt]
  & Subcutaneous Emphysema & 716 \\
  & Degenerative Changes & 712 \\
  & Hilar Abnormality & 431 \\
  & COPD/Emphysema & 358 \\
  & Heart Failure & 357 \\
  & Aortic Abnormality & 354 \\
  & Tracheal Deviation & 352 \\
  & Scoliosis & 241 \\
  & Osteopenia/Osteoporosis & 228 \\
  & Kyphosis & 141 \\
  & Bronchiectasis & 83 \\
  & Pneumomediastinum & 82 \\
  & Pneumoperitoneum & 80 \\
  & Hiatal Hernia & 64 \\
  & Elevated Hemidiaphragm & 48 \\
  & Pericardial Effusion & 46 \\
  & Lymphadenopathy & 41 \\
\midrule
\multicolumn{2}{@{}l}{\textbf{Subtotal unmapped (17 findings)}} & \textbf{4,334 (3.4\%)} \\
\midrule
\multicolumn{2}{@{}l}{\textbf{Grand total (59 findings)}} & \textbf{126,231} \\
\bottomrule
\end{tabular}
\end{table}

\begin{figure}[H]
\centering
\makebox[\textwidth][c]{\includegraphics[width=1.3\textwidth]{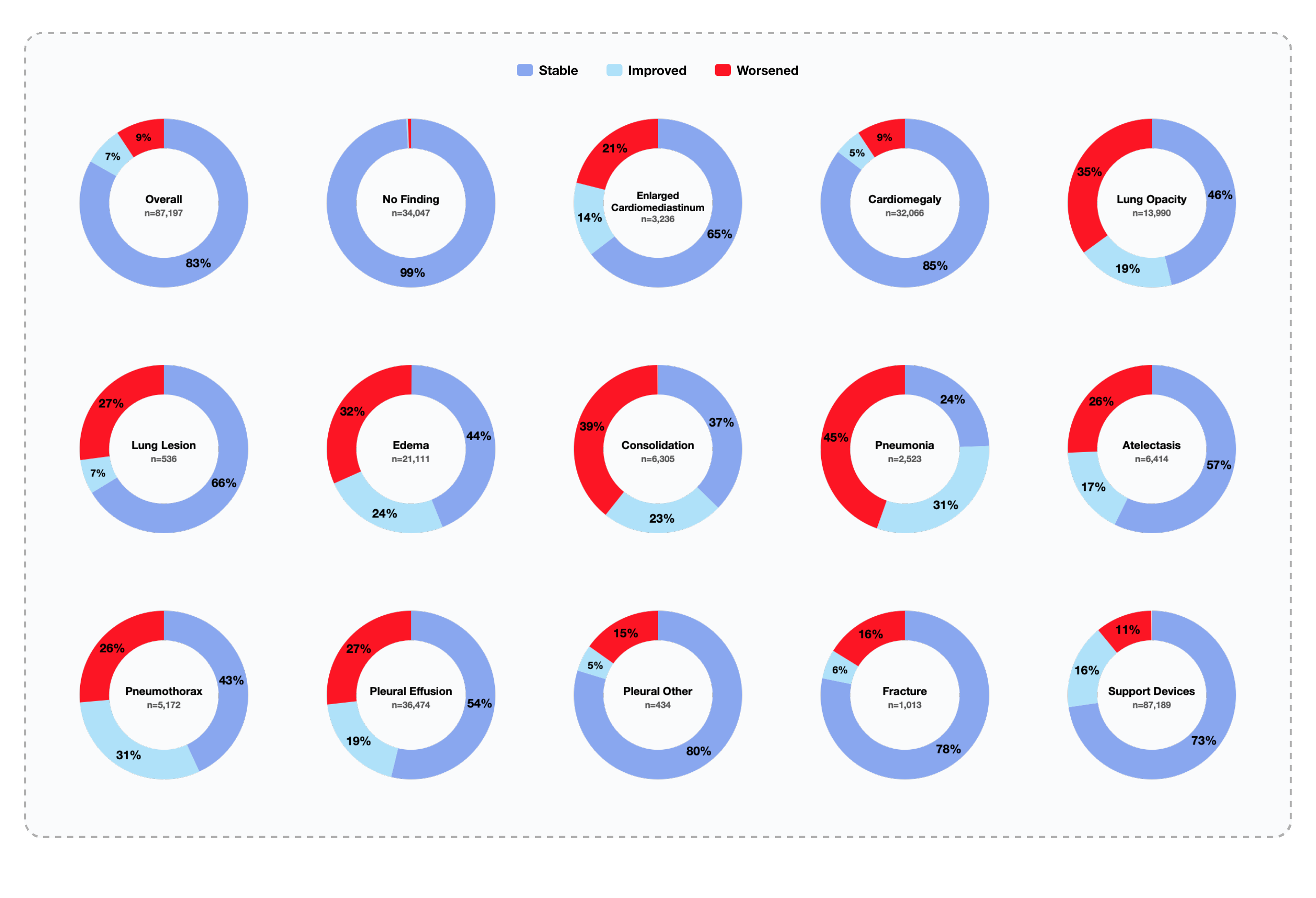}}
\caption{Temporal trajectory distribution within chains-of-thought for patients with serial studies.}
\label{fig:Temporal}
\end{figure}

\begin{table}[H]
\centering
\caption{Within-model normal-vs-pathology comparison. Significance from Mann--Whitney $U$ tests.}
\label{tab:normvspath}
\scriptsize
\setlength{\tabcolsep}{4pt}
\renewcommand{\arraystretch}{1.15}
\resizebox{\textwidth}{!}{%
\begin{tabular}{@{}llcccc@{}}
\toprule
\textbf{Dimension} & \textbf{Model} & \textbf{Normal} & \textbf{Pathology} & \textbf{$\Delta$ (P$-$N)} & \textbf{$p$-value} \\
\midrule
\multirow{4}{*}{Comprehensiveness}
 & CheXthought Data       & 4.63 (4.56--4.70) & 4.70 (4.66--4.75) & $+$0.07 & $p = 0.60$ \\
 & GPT~5.2            & 4.24 (4.16--4.31) & 4.25 (4.18--4.32) & $+$0.01 & $p = 0.86$ \\
 & MedGemma~1.5     & 4.04 (3.99--4.10) & 4.28 (4.21--4.35) & $+$0.24 & $p < 0.00001$ \\
 & Claude Opus~4.5    & 4.07 (3.99--4.14) & 4.00 (3.93--4.07) & $-$0.06 & $p = 0.60$ \\
\midrule
\multirow{4}{*}{Causal Support}
 & CheXthought Data       & 4.80 (4.75--4.84) & 4.87 (4.84--4.90) & $+$0.07 & $p = 0.033$ \\
 & GPT~5.2            & 4.20 (4.12--4.28) & 4.51 (4.45--4.57) & $+$0.31 & $p < 0.000001$ \\
 & MedGemma~1.5     & 3.89 (3.81--3.97) & 4.07 (3.97--4.17) & $+$0.18 & $p = 0.014$ \\
 & Claude Opus~4.5    & 4.10 (4.02--4.18) & 4.38 (4.31--4.45) & $+$0.28 & $p < 0.00001$ \\
\midrule
\multirow{4}{*}{Factuality}
 & CheXthought Data      & 4.74 (4.69--4.79) & 5.00 (5.00--5.00) & $+$0.26 & $p < 0.0001$ \\
 & GPT~5.2            & 4.10 (4.05--4.15) & 4.26 (4.19--4.33) & $+$0.16 & $p < 0.001$ \\
 & MedGemma~1.5     & 3.74 (3.61--3.87) & 4.47 (4.39--4.54) & $+$0.73 & $p < 0.0001$ \\
 & Claude Opus~4.5    & 3.73 (3.63--3.83) & 3.92 (3.82--4.01) & $+$0.19 & $p = 0.006$ \\
\midrule
\multirow{4}{*}{Spatial Localization}
 & CheXthought Data     & 4.78 (4.73--4.83) & 4.97 (4.95--4.99) & $+$0.19 & $p < 0.0001$ \\
 & GPT~5.2            & 4.20 (4.10--4.28) & 3.42 (3.30--3.54) & $-$0.78 & $p < 0.0001$ \\
 & MedGemma~1.5     & 4.03 (3.96--4.10) & 4.62 (4.56--4.68) & $+$0.59 & $p < 0.0001$ \\
 & Claude Opus~4.5    & 4.07 (3.96--4.17) & 2.96 (2.86--3.06) & $-$1.11 & $p < 0.0001$ \\
\bottomrule
\end{tabular}%
}
\end{table}

\begin{table}[H]
\centering
\caption{Per-pathology detection rate (\%) with 95\% bootstrap confidence intervals on NIH ChestX-ray14 and MIMIC-CXR. Best result per row in \textbf{bold}. $^{\dagger}$Significantly outperforms $\geq$3 other models; $^{*}$significantly outperforms $\geq$1 other model (McNemar's test, $p<0.05$).}
\label{tab:PathologyExtendedData}
\scriptsize
\setlength{\tabcolsep}{3.5pt}
\renewcommand{\arraystretch}{1.15}
\resizebox{\textwidth}{!}{%
\begin{tabular}{@{}ll cccccc@{}}
\toprule
\textbf{Dataset} & \textbf{Pathology} & \textbf{CheXthought-VLM} & \textbf{CheXthought-CoT} & \textbf{Synthetic-CoT} & \textbf{CheXpert-Report} & \textbf{Qwen3-VL-8B-Thinking} & \textbf{MedGemma~1.5} \\
\midrule
NIH & Atelectasis        & \textbf{81 (65--96)}$^{\dagger}$ & 38 (19--58)$^{*}$ & 19 (4--35) & 65 (46--85)$^{*}$ & 0 (0--0) & 69 (50--85)$^{*}$ \\
NIH & Cardiomegaly       & 75 (58--92)$^{*}$ & \textbf{92 (79--100)}$^{\dagger}$ & 4 (0--12) & 42 (21--62)$^{*}$ & 12 (0--25) & 50 (29--71)$^{*}$ \\
NIH & Edema              & 42 (21--62)$^{*}$ & \textbf{67 (46--83)}$^{\dagger}$ & 29 (12--50) & 58 (38--79)$^{*}$ & 4 (0--12) & 42 (21--62)$^{*}$ \\
NIH & Pleural Effusion   & \textbf{84 (68--96)}$^{\dagger}$ & 72 (52--88)$^{*}$ & 20 (4--36) & 68 (48--84)$^{*}$ & 28 (12--48) & 80 (64--96)$^{*}$ \\
NIH & Pneumonia          & \textbf{28 (12--48)}$^{*}$ & 12 (0--24) & 8 (0--20) & 20 (8--36) & 0 (0--0) & 12 (0--24) \\
NIH & Pneumothorax       & 46 (27--65)$^{*}$ & 38 (19--58)$^{*}$ & 8 (0--19) & \textbf{50 (31--69)}$^{*}$ & 0 (0--0) & 35 (15--54)$^{*}$ \\
\midrule
MIMIC & Atelectasis      & 65 (46--85)$^{\dagger}$ & 38 (19--58)$^{*}$ & 4 (0--12) & \textbf{81 (65--96)}$^{\dagger}$ & 0 (0--0) & 23 (8--38)$^{*}$ \\
MIMIC & Cardiomegaly     & 62 (42--79)$^{*}$ & 50 (29--71)$^{*}$ & 33 (17--54)$^{*}$ & 33 (17--54)$^{*}$ & 0 (0--0) & \textbf{88 (71--100)}$^{\dagger}$ \\
MIMIC & Edema            & 42 (25--62)$^{*}$ & 46 (25--67)$^{*}$ & 8 (0--21) & 0 (0--0) & 0 (0--0) & \textbf{100 (100--100)}$^{\dagger}$ \\
MIMIC & Pleural Effusion & 88 (76--100)$^{\dagger}$ & 56 (36--76)$^{*}$ & 44 (24--64)$^{*}$ & 80 (64--96)$^{\dagger}$ & 8 (0--20) & \textbf{96 (88--100)}$^{\dagger}$ \\
MIMIC & Pneumonia        & 4 (0--12) & 40 (20--60)$^{*}$ & 20 (4--36) & \textbf{80 (64--96)}$^{\dagger}$ & 0 (0--0) & 28 (12--48)$^{*}$ \\
MIMIC & Pneumothorax     & 35 (15--54)$^{*}$ & 42 (23--62)$^{*}$ & 8 (0--19) & 38 (19--58)$^{*}$ & 0 (0--0) & \textbf{88 (73--100)}$^{\dagger}$ \\
\bottomrule
\end{tabular}%
}
\end{table}

\section{Supplementary information}

\setcounter{figure}{0}
\setcounter{table}{0}
\renewcommand{\figurename}{Supplementary Figure}
\renewcommand{\tablename}{Supplementary Table}

\begin{figure}[H]
\centering
\includegraphics[width=\textwidth]{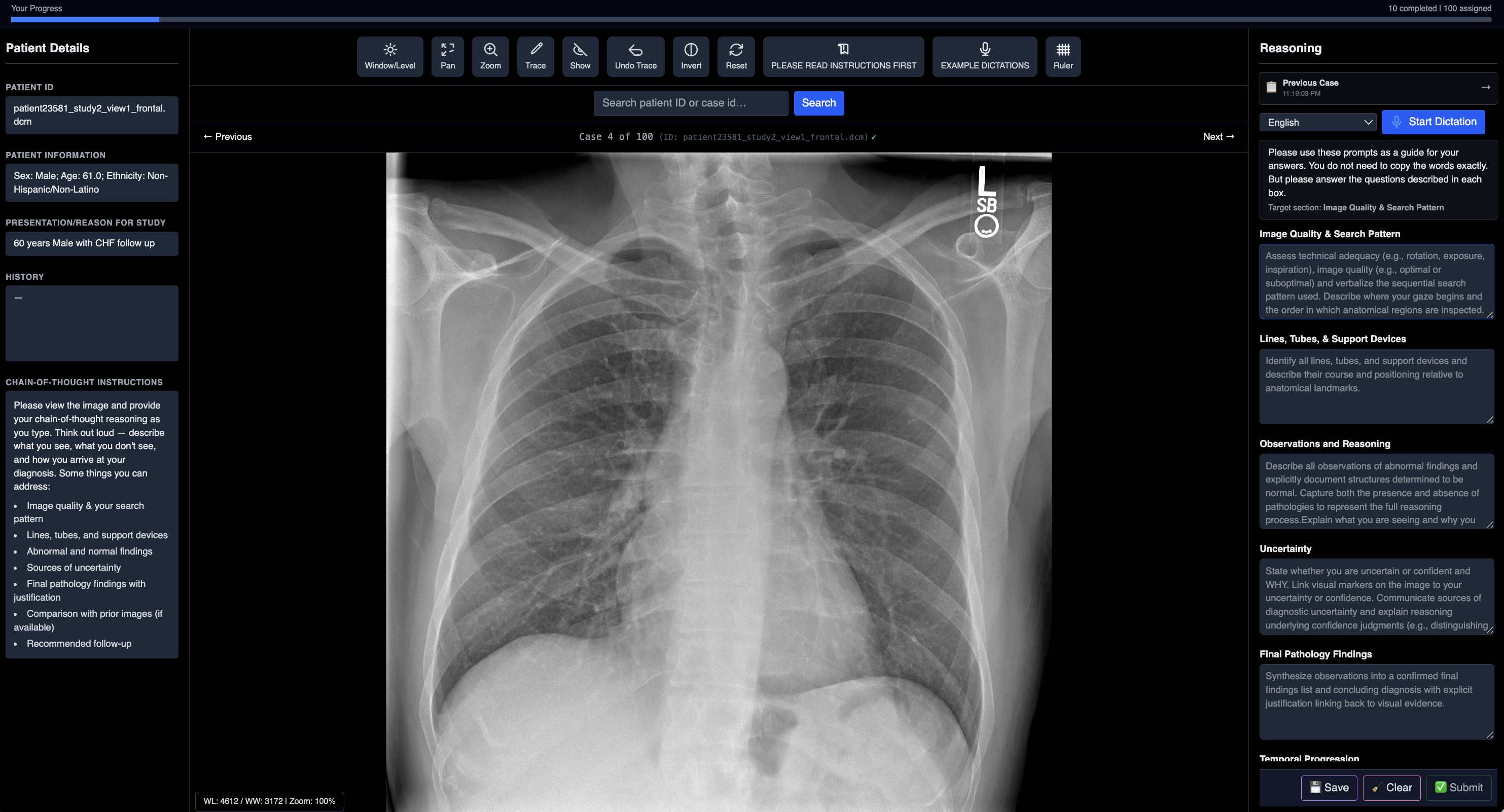}
\caption{
Screenshot of the web-based DICOM viewer annotation platform used to collect chain-of-thought reasoning and visual attention annotations during chest X-ray interpretation.
}
\label{fig:Supplement1}
\end{figure}

\begin{table}[H]
\caption{Structured chain-of-thought annotation framework}
\label{tab:CoT}
\footnotesize
\renewcommand{\arraystretch}{1.3}
\begin{tabularx}{\textwidth}{@{}>{\raggedright\arraybackslash\hyphenpenalty=10000}p{2.8cm}X@{}}
\toprule
\textbf{Component} & \textbf{Description} \\
\midrule
\textbf{Image Quality \& Search Pattern} &
Assess technical adequacy (rotation, exposure, inspiration) and image quality, and verbalize the sequential search pattern, including where visual attention and eye-path begins and the order in which anatomical regions are inspected. \\
\midrule
\textbf{Lines, Tubes, Support Devices} &
Identify all lines, tubes, and support devices and describe their course and positioning relative to anatomical landmarks. \\
\midrule
\textbf{Observations} &
Describe all abnormal findings and explicitly document structures determined to be normal, capturing both the presence and absence of pathologies. \\
\midrule
\textbf{Uncertainty} &
Communicate sources of diagnostic uncertainty and the reasoning underlying confidence judgements (e.g., distinguishing consolidation from atelectasis, or identifying pleural effusion via the meniscus sign). \\
\midrule
\textbf{Final Pathology Findings} &
Synthesise observations into a final findings list and concluding diagnosis, with explicit justification linking back to visual evidence. \\
\midrule
\textbf{Temporal Comparison of Serial Images} &
Compare with prior examinations and document interval changes (improvement, worsening, stability). \\
\midrule
\textbf{Recommended Follow-up} &
Suggest additional imaging, testing, or clinical actions to resolve residual diagnostic uncertainty. \\
\bottomrule
\end{tabularx}
\end{table}

\begin{table}[H]
\centering
\caption{Evaluation rubric for chain-of-thought assessment}
\label{tab:rubric}
\footnotesize
\setlength{\tabcolsep}{6pt}
\renewcommand{\arraystretch}{1.1}

\begin{tabular}{p{3.5cm} p{11cm}}
\toprule
\textbf{Dimension} & \textbf{Scoring criteria} \\
\midrule

\textbf{Comprehensiveness of Findings} &
Accuracy and quantity of pathological findings identified in the image.
\textbf{1:} $<$25\% identified; major errors.
\textbf{2:} 25--49\% identified; key findings missed.
\textbf{3:} 50--74\% identified; incomplete.
\textbf{4:} 75--90\% identified; minor error.
\textbf{5:} $>$90\% identified with no meaningful errors. \\

\midrule

\textbf{Causal Support} &
Whether final findings are supported by reasoning.
\textbf{1:} Unsupported findings.
\textbf{2:} Partial support.
\textbf{3:} Mostly supported but vague.
\textbf{4:} Nearly all supported.
\textbf{5:} Fully supported with explicit evidence. \\

\midrule

\textbf{Factuality} &
Absence of hallucinated findings.
\textbf{1:} $>$5 hallucinations.
\textbf{2:} 3--4 hallucinations.
\textbf{3:} 1--2 hallucinations.
\textbf{4:} One minor hallucination.
\textbf{5:} No hallucinations. \\

\midrule

\textbf{Spatial Localization} &
Correct anatomical and laterality description.
\textbf{1:} Mostly incorrect.
\textbf{2:} Frequent errors.
\textbf{3:} $\sim$50\% correct.
\textbf{4:} Mostly correct.
\textbf{5:} Fully accurate localization. \\

\bottomrule
\end{tabular}
\end{table}
\end{document}